\documentclass[times, review, 10pt]{elsarticle}

\usepackage{amssymb}
\usepackage{amsmath}

\usepackage{subcaption}
\usepackage{algorithm}
\usepackage{algorithmic}

\expandafter\let\csname equation*\endcsname\relax
\expandafter\let\csname endequation*\endcsname\relax
\usepackage{amssymb}
\usepackage{amsmath}

\usepackage{booktabs}
\usepackage{multirow}
\usepackage{siunitx}  
\expandafter\let\csname equation*\endcsname\relax
\expandafter\let\csname endequation*\endcsname\relax
\usepackage{amssymb}
\usepackage{amsmath}

\usepackage{pifont}

\newcommand{\hz}[1]{}

\journal{Nuclear Physics B}

\begin{document}

\begin{frontmatter}




\title{FreSia: Frequency-Semantic Instantiation and Alignment for Multivariate Time Series Analysis}


\author[label1]{Yubo Wang} 
\ead{wangyb@bit.edu.cn}
\author[label2]{Hui He} 
\author[label2]{Hezhe Qiao} 
\author[label1]{Guoqing Ji} 
\author[label1]{Zhendong Niu\corref{cor1}} 
\ead{zniu@bit.edu.cn}
\affiliation[label1]{organization={School of Computer Science and Technology, Beijing Institute of Technology},
            addressline={No.5 Zhongguancun South Street}, 
            city={Beijing},
            postcode={100081}, 
            country={China}}
\affiliation[label2]{organization={School of Computing and Information Systems, Singapore Management University},
            addressline={80 Stamford Road}, 
            postcode={178902}, 
            country={Singapore}}
\cortext[cor1]{Corresponding author}

\begin{abstract}
 {Large Language Models (LLMs) have shown strong potential in multivariate time series forecasting and anomaly detection. Existing studies predominantly inject temporal information into LLMs via direct numerical tokenization or heuristic textual descriptions. However, LLMs still face difficulty in perceiving the underlying structural patterns of numerical time series, particularly the seasonal and trend components obscured by discrete numerical tokens. To bridge this gap, we propose \textbf{FreSia}, a frequency-aware framework that establishes an effective alignment between the semantic space of LLMs and the frequency space of time series. Specifically, \textbf{FGPrompt}, a \underline{F}requency-\underline{G}uided \underline{Prompt} mechanism within FreSia, distills the frequency-domain structures of time series and projects them into prompts tailored to the semantic space of LLMs. Furthermore, we introduce a Global-driven Context Learning (\textbf{GCL}) component, which uses a global CLS-driven probe to generate global context to bridge the time-frequency domain gap and fuse the multi-modal information. Experiments on eight forecasting benchmarks show that FreSia achieves average improvements of 13.48\% and 8.06\% in MSE and MAE, respectively. }

\end{abstract}




\end{frontmatter}




\begin{figure}[h]
\centering
\subfloat[Representation gap]
{\includegraphics[width=1.0\columnwidth]{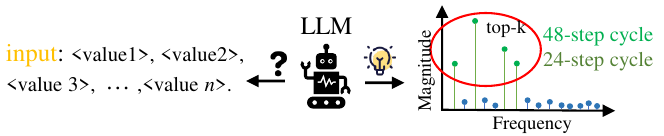}}\\

\subfloat[Spectral sparsity in real-world time series]
{\includegraphics[width=1.0\columnwidth]{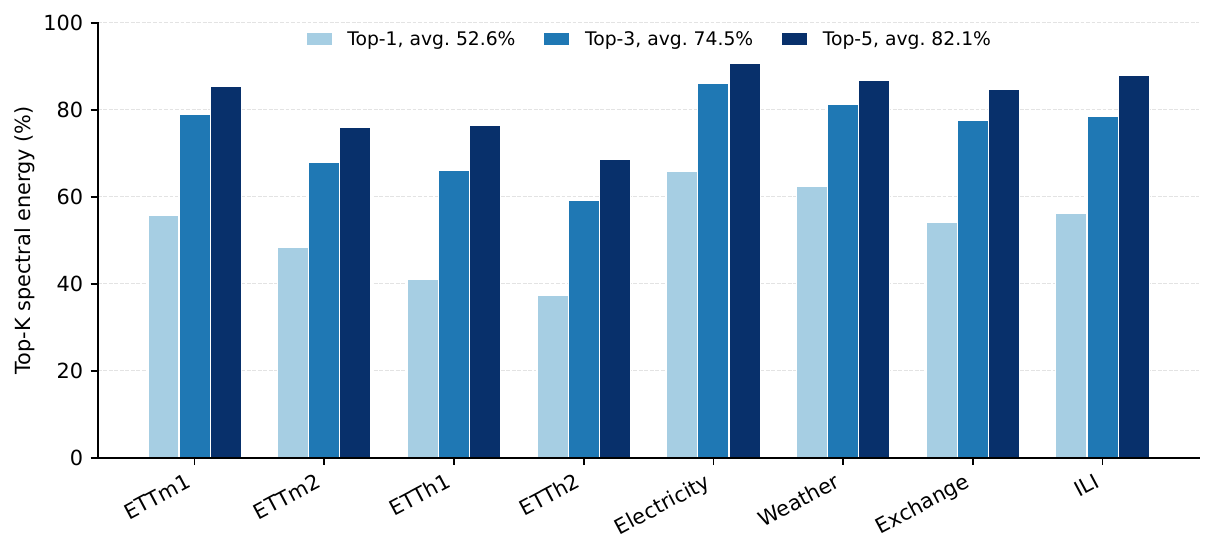}}
\caption{ {(a) Representation gap between raw numerical prompts and frequency-guided semantic prompts. (b) Top-K frequency components explain a large portion of spectral energy across eight benchmark datasets. This motivates the use of dominant frequency components as compact structural primitives for frequency-guided semantic instantiation.}}
\label{fig1}
\end{figure}

\section{Introduction}
\hz{Double check the extra space between test and [Ref], some spaces are not only one}

Multivariate time series analysis plays a pivotal role in a wide range of real-world applications~\cite{wen2022robust}, such as traffic flow forecasting~\cite{fang2025efficient} and industrial anomaly detection~\cite{pemula2025robust}. With the rapid development of NLP, pre-trained LLMs have been widely applied to time series analysis. In particular, prompt-based LLM methods introduce textual prompts as additional inputs to aid LLMs in understanding temporal tasks. Such prompts typically contextualize the statistical features of time series with textual descriptions: either by converting data into textual representations~\cite{liu2025timecma} or by directly summarizing sequence information using natural language~\cite{zhou2023one}. Subsequently, the prompts are encoded by a pre-trained LLM to obtain robust semantic representations for downstream tasks.



Existing LLM-based time series methods mainly inject temporal information through raw numerical serialization or time-domain statistical descriptions. For instance, Time-LLM~\cite{ICLR2024_680b2a81} integrates domain knowledge and input statistics into descriptive prompts, which are then concatenated with temporal embeddings. TimeCMA~\cite{liu2025timecma} generates prompts based on historical values and trends, utilizing them as semantic anchors for cross-modal retrieval. Similarly, LangTime~\cite{pmlr-v267-niu25e} treats time series directly as language tokens to leverage pretrained LLMs. Although these strategies make numerical data accessible to LLMs, they still suffer from an important representation mismatch. 

 As illustrated in Figure~\ref{fig1} (a), raw numerical values are dense, continuous, scale-sensitive, and tokenization-unfriendly. LLMs are primarily pretrained on discrete linguistic tokens and are effective at modeling semantic and logical relations in natural language, but such token-based representations are not naturally aligned with continuous numerical dynamics~\cite{ICLR2024_680b2a81}. When a time series is serialized into a long sequence of numerical tokens, its underlying temporal structures, such as periodicity and seasonality, remain implicit. Frequency-domain analysis provides a structured intermediate representation by exposing periodic and harmonic components. These components can be further discretized and verbalized as symbolic cycle descriptors, such as daily cycle, half-day cycle, weekly cycle, or general x-step cycle, making them more compatible with the token-based semantic representation space of LLMs. We therefore select dominant frequency components based on their spectral energy. According to the energy-preserving property of the discrete Fourier transform, the DFT decomposes signal energy across orthogonal frequency components, making high-energy components natural candidates for capturing the main temporal variations.

 As shown in Figure~\ref{fig1} (b), the Top-3 and Top-5 dominant frequencies explain 74.5\% and 82.1\% of non-DC spectral energy on average across eight benchmark datasets. This observation suggests that dense numerical sequences can be compactly summarized by a few dominant spectral primitives. It motivates us to express dominant frequency-domain structures through natural language descriptors, which provide a compact semantic abstraction of the main temporal patterns. A detailed explanation of the energy-preserving property of DFT is provided in ~\ref{app:energy-preserving}. 



Motivated by this observation, we propose \textbf{FreSia}, a framework for MTS analysis through frequency-guided prompting and global-driven contextualization. Our approach consists of two key components: \textbf{(1) Frequency-Guided Prompt (FGPrompt):} To bridge the  {representation} gap between raw numbers and semantic reasoning, we distill the \textit{intrinsic frequency-domain structure} of time series using FFT. By mapping Top-$K$ frequencies into the Frequency Pool, we transform numerical observations into structured  {spectral} primitives. \textbf{(2) Global-driven Context Learning (GCL):} To bridge the representation gap between frequency-domain structure and time-domain contextual dynamics, we introduce a Global-driven Context Learning mechanism. A global CLS-driven probe is employed to capture the context from the temporal encoder. By fusing this temporal context  with spectral primitives, we can bridge time-domain and frequency-domain representations.



\noindent Our contributions are summarized as follows:
\begin{itemize}
\item We develop an FGPrompt to distill intrinsic spectral structures into a learnable frequency pool. This enables the LLM to retrieve frequency-domain structures rather than relying solely on numerical summary statistics.
\item We propose a GCL architecture where a CLS-driven probe generates instance-specific residual offsets to calibrate the prompt space. This design bridges the gap between the time and frequency domains.
\item We evaluate FreSia on both multivariate time-series forecasting and reconstruction-based anomaly detection. With task-specific heads, FreSia achieves competitive results across the evaluated forecasting and anomaly-detection benchmarks.
\end{itemize}

\section{Related work}
\subsection{Deep Time Series Analysis}
Deep learning provides an effective way to model multivariate time series by learning temporal dependencies. TimesNet~\cite{DBLP:conf/iclr/WuHLZ0L23} captures multi-period patterns through temporal 2D-variations. PatchTST~\cite{nie2023iclr-time} models long-range dependencies with patch-based representations, while iTransformer~\cite{liu2024iclr-itransformer} captures inter-variable dependencies by multivariate-attention. Recently, SEMPO~\cite{DBLP:conf/nips/HeYMZNP25} further explores lightweight foundation modeling for time-series forecasting. Frequency-domain analysis has received increasing attention in time-series forecasting. FEDformer~\cite{zhou2022fedformer} combines seasonal-trend decomposition with frequency-enhanced modeling to capture temporal patterns in the frequency domain. FredFormer~\cite{piao2024fredformer} addresses frequency bias, FilterNet~\cite{yi2024filternet} learns temporal patterns through frequency filtering, and ReFocus~\cite{DBLP:journals/corr/abs-2502-16890} emphasizes mid- and key-frequency components. TimeKAN~\cite{huang2025timekan} combines frequency decomposition with multi-order KANs, capturing nonlinear dynamics across different frequency bands. FDNet~\cite{DBLP:journals/pr/HuWDDWHWX26} disentangles informative high-frequency components from noise through information-theoretic objectives, aiming to preserve useful short-term dynamics rather than indiscriminately suppressing high-frequency information. MTCMD~\cite{zhang2026multi} models long- and short-term cyclical variations in a multi-dimensional temporal space and extracts seasonal and trend components across multiple scales. Despite these advances, these methods primarily operate in the numerical representation space. In anomaly detection, AdaCLIP~\cite{DBLP:conf/eccv/CaoZFCSB24} adapts CLIP with hybrid learnable prompts, while Cao et al.~\cite{DBLP:journals/tcyb/CaoXCSDGS25} further personalize vision-language models for zero-shot anomaly detection. More recently, a large-scale benchmark~\cite{DBLP:journals/pr/CaoCZXZSTHHS26} was introduced to evaluate visual anomaly detection under complex view-illumination conditions. In contrast, FreSia transforms dominant spectral structures into semantic cycle descriptors and aligns them with the representation space of LLMs. Another key task in time series analysis is anomaly detection. General time-series architectures can also be adapted to anomaly detection. FPT~\cite{zhou2023one} leverages pretrained language models for general time-series representation learning, while TimesNet~\cite{DBLP:conf/iclr/WuHLZ0L23} captures multi-period temporal variations and PatchTST~\cite{nie2023iclr-time} models temporal dependencies through patch-based representations. These models can be employed in reconstruction-based detection settings. In addition, there are dedicated models specifically designed for anomaly detection tasks. AnomalyTransformer~\cite{xu2022iclr-anomaly} characterizes anomalies through association discrepancy. GCAD~\cite{liu2025gcad} combines graph modeling with Granger Causality to capture inter-variable relationships for anomaly detection. TimeRadar~\cite{DBLP:conf/kdd/HeQCYP26} introduces a domain-rotatable foundation model for time-series anomaly detection. Recently, researchers have explored combining Mixture of Experts (MoE) with graph structures. Graph-MoE~\cite{huang2025graph} integrates MoE and graph learning to improve representation ability and anomaly detection performance.



\subsection{LLM-empowered Time Series Analysis}
The success of LLMs has sparked significant interest in adapting their reasoning capabilities. $\mathrm{S^{2}IP}$-LLM~\cite{pan2024s} aligns time-series representations with the semantic space of LLMs through prompt learning, while LLM4TS~\cite{chang2025llm4ts} adapts pre-trained LLMs for time-series forecasting through parameter-efficient alignment. Another models focus on prompt-based adaptation, which guides LLMs with textual descriptions. For instance, ChatTime~\cite{wang2025chattime} integrates processed temporal sequences with textual prompts as inputs to LLMs, while the model~\cite{xiong2025beyond} leverages LLMs to learn textual representations of temporal patterns for forecasting. Furthermore, Time-LLM~\cite{ICLR2024_680b2a81} incorporates domain knowledge and input statistics, while TimeCMA~\cite{liu2025timecma} and LLM-TPF~\cite{pan2025llm} utilize historical trends and numerical extrema as instructions. Achieving seamless integration between numerical features and semantic prompts is critical for cross-modal forecasting. Existing studies adopt different strategies to integrate temporal and semantic representations. UniTime~\cite{liu2024unitime} uses domain instructions and a Language-TS Transformer for cross-modal alignment, GPT4MTS~\cite{jia2024gpt4mts} jointly incorporates numerical and textual information, and STG-LLM~\cite{DBLP:journals/corr/abs-2401-14192} employs a tokenizer-adapter architecture to bridge structured temporal data and LLMs. To facilitate more efficient temporal-textual exchange, ITFormer~\cite{pmlr-v267-wang25av} bridges time series data with frozen LLMs via a dedicated interaction module, while Context-Alignment~\cite{hu2025context} further investigates the alignment between temporal contexts and LLM representations to better activate pretrained semantic knowledge for time-series modeling. More recent studies explore richer forms of contextual integration. TimeCMA~\cite{liu2025timecma} retrieves time-series information from prompt representations through cross-modality alignment, while TimeCAP~\cite{lee2025timecap} employs LLM agents to contextualize time-series data for event prediction.



LLMs have also been explored for time-series anomaly detection. LLMAD~\cite{liu2025large} leverages LLMs to provide interpretable anomaly analysis by combining detection results with semantic explanations, while TSINR~\cite{li2025tsinr} reconstructs temporal observations and identifies anomalies according to reconstruction discrepancies. These studies approach anomaly detection through semantic interpretation or reconstruction-based modeling. In contrast, FreSia incorporates frequency-derived structural information into semantic representations for reconstruction-based anomaly detection.

\section{Preliminaries}
\paragraph{Semantic Prompt Induction} 
Given a multivariate time series $\mathbf{X}_L$, we derive a set of semantic instructions $\mathcal{I} = \{I_1, I_2, \dots, I_C\}\in \mathbb{R}^{C \times S}$, where each $I_i$ corresponds to the $i$-th variable (channel) and has $S$ elements. 

\paragraph{Multivariate Time Series Forecasting}
 {
Formally, MTS forecasting uses the lookback window $\mathbf{X}_{L} = \{ \mathbf{x}_{t-L+1}, \mathbf{x}_{t-L+2}, $ $\dots, \mathbf{x}_{t} \} \in \mathbb{R}^{C \times L}$ to predict the future sequence $\mathcal{Y}_{T} = \{ \mathbf{x}_{t+1}, \mathbf{x}_{t+2}, \dots, \mathbf{x}_{t+T} \} \in \mathbb{R}^{C \times T}$ with $\mathcal{I}$. $C$ is the number of variables and $L$ represents the window length. The $\mathbf{x}_t \in \mathbb{R}^C$ represents the observed multivariate data at $t$. 
}


\paragraph{Multivariate Time Series Anomaly Detection} 
Given an input $\mathbf{X}_L \in \mathbb{R}^{C \times L}$, the objective is to predict a binary sequence $\mathcal{A} = \{a_1, \dots, a_L\} \in \{0, 1\}^L$. Here, $a_i = 1$ denotes an anomaly at timestamp $i$, while $a_i = 0$ represents normal data. We adopt a reconstruction-based approach where the model generates $\hat{\mathbf{X}}_L = f(\mathbf{X}_L, \mathcal{I})$. An anomaly is identified if the reconstruction error $s_i = \|\mathbf{x}_i - \hat{\mathbf{x}}_i\|^2$ exceeds a specific threshold.


\begin{figure*}[!t]
\centering
\includegraphics[width=1.0\textwidth]{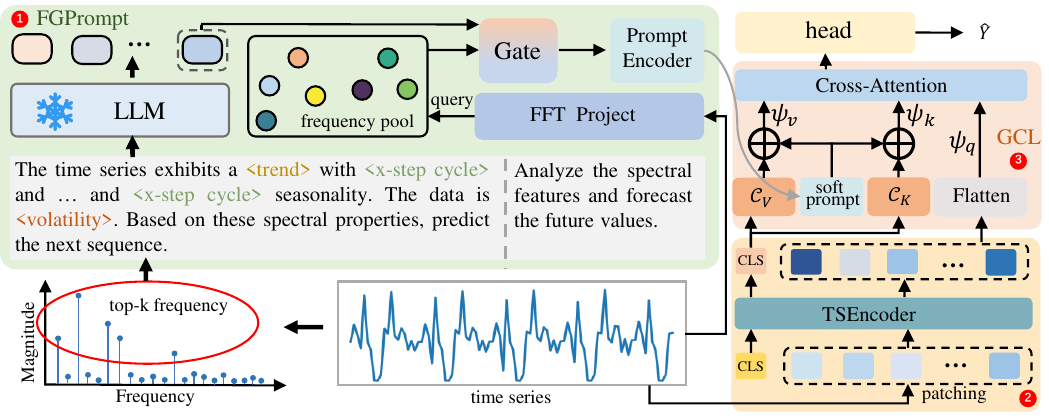}
\caption{Framework of \textbf{FreSia}. \textcolor{red}{\ding{182}} Frequency-Guided Prompt (FGPrompt) injects frequency-domain information into prompts and the frequency pool. \textcolor{red}{\ding{183}} TSEncoder captures temporal dependencies and generates a CLS representation. \textcolor{red}{\ding{184}} Global-driven Context Learning (GCL) bridges the time-frequency domain gap and fuses multi-modal information.}
\label{fig2}
\end{figure*}

\section{Methodology}
\paragraph{Framework Overview}
The proposed framework consists of three components: (1) \textbf{FGPrompt}, (2) \textbf{TSEncoder}, (3) \textbf{Global-driven Context Learning}. Specifically, we leverage a frequency-driven instruction module to generate prompts. Then, the generated prompt is fused with a latent frequency pool through a gating mechanism. Concurrently, the time-series encoder extracts temporal features and global representations with a CLS token, and we use the learned CLS representation to calibrate the prompt embeddings. These multimodal representations are fused through an alignment module to forecast.

\subsection{Frequency-Guided Prompt}
To bridge the gap between numerical time series and the semantic space of LLMs, we propose a Frequency-Guided Prompt (FGPrompt). This module extracts time-domain and frequency-domain information from the time series to form a prompt. In this way, we can provide the model with explicit prior information about the signal's frequency characteristics.

\begin{table}[htbp]
\centering
\small
\begin{tabular}{l|c|l}
\toprule
\textbf{Property} & \textbf{Numerical Range} & \textbf{Linguistic Descriptor} \\
\midrule
\multirow{5}{*}{Trend ($\hat{k}$)} & $\hat{k} > 0.5$ & strong upward trend \\
 & $0.1 < \hat{k} \le 0.5$ & slight upward trend \\
 & $-0.1 \le \hat{k} \le 0.1$ & stable trend \\
 & $-0.5 \le \hat{k} < -0.1$ & slight downward trend \\
 & $\hat{k} < -0.5$ & strong downward trend \\
\midrule
\multirow{3}{*}{Volatility ($v$)} & $v > 1.0$ & high volatility \\
 & $0.5 < v \le 1.0$ & moderate volatility \\
 & $v \le 0.5$ & smooth \\
\bottomrule
\end{tabular}
\caption{Mapping rules for linguistic descriptors in Frequency-Guided Prompts.}
\label{table0}
\end{table}

\paragraph{Feature Characterization}
{
For each input sequence $\mathbf{x}$, FGPrompt extracts compact semantic descriptors from both frequency-domain and time-domain information. The frequency-derived cycle descriptors serve as the primary structural primitives, while trend and volatility descriptors provide auxiliary contextual cues. Specifically, dominant frequencies are identified by applying FFT to $\mathbf{x}$, and their corresponding periods $T_i = 1/f_i$ are mapped to semantic cycle descriptors according to the sampling frequency, such as daily, weekly, or monthly cycles. Here, \(f_i\) denotes the frequency in cycles per time step. In addition, a linear fit $y = kt + b$ is used to estimate the global trend, and the coefficient of variation of first-order differences is used to characterize local volatility. These numerical features are then converted into linguistic descriptors, with the detailed quantization rules summarized in Table~\ref{table0}.
}

\paragraph{Semantic Prompt Formulation and Encoding} The extracted features are aggregated into a structured linguistic Prompt $I$, which follows a template: "The time series exhibits a [Trend] with significant [Seasonality] seasonality. The data is [Volatility]. Based on these properties, predict the next sequence." To project these semantic prompts into the latent space, we use pre-trained GPT-2 as a feature extractor. Let $\text{Tokenizer}(\cdot)$ and $\text{Enc}(\cdot)$ denote the LLM's tokenizer and transformer backbone, respectively. The semantic embedding $\mathbf{E}_{LLM}$ is derived from the hidden state of the last token:
\begin{equation}
    \mathbf{E}_{LLM} = \text{Enc}(\text{Tokenizer}(I)) 
\end{equation}


\paragraph{ Frequency Pool} To bridge the gap between discrete linguistic tokens and continuous numerical features, we introduce a learnable frequency pool $\mathbf{P}=\{p_{1},...,p_{M}\} \in \mathbb{R}^{M \times d}$. This pool represents the structural archetypes of time-series dynamics that can be retrieved using frequency amplitude and phase features. We use the frequency amplitude and phase features $\mathbf{f}_{x}$ as a query to retrieve relevant expertise from the pool:
\begin{equation}
\mathbf{P}_{freq}= \rm Ret(\phi(\mathbf{f}_{x}), \mathbf{P}), \quad
\gamma_{b}=\sigma \left( g([\mathbf{P}_{freq}||\mathbf{E}_{LLM}]) \right ) 
\end{equation}
Among these, $\phi$ is a linear mapping, and $\rm Ret$ denotes retrieving relevant vectors from the frequency pool via weighted methods. We concatenate the query results with the LLM-embedded prompt words, then pass this through a gated network g composed of a lightweight MLP to obtain the fusion weights $\gamma_{b}$. These are subsequently fused into $\mathcal{I}_{soft}$.
\begin{equation}
\mathcal{I}_{soft}= \gamma_{b}\mathbf{E}_{LLM}+(1-\gamma_{b})\mathbf{P}_{freq}
\end{equation}
Then, we use a transformer-based $\text{PromptEncoder}(\cdot)$ to process the semantic embedding $\mathcal{I}_{soft}$. The final semantic embedding is obtained as:
\begin{equation}
    \mathcal{I}_{soft}= \text{PromptEncoder}(\mathcal{I}_{soft})
\end{equation}

\subsection{TSEncoder}
To extract temporal dependencies and global representations from time series data, we introduce the TSEncoder. We first employ a patching strategy to process normalized time series. Subsequently, we embed the patch representations. We prepend a learnable CLS token $\mathbf{z}_{CLS}$ to the patch embeddings $\mathbf{E}_{seq}$ to aggregate global information from the input sequence. Unlike the localized patch embeddings that encode local temporal variations, the learned CLS representation summarizes information across the entire look-back window. The concatenated sequence $[\mathbf{z}_{CLS}; \mathbf{E}_{seq}]$ is then processed by the Transformer-based TSEncoder:

\begin{equation}
[\mathbf{E}_{\mathrm{Time}}^{\mathrm{CLS}}; \mathbf{E}_{\mathrm{Time}}]
=
\mathrm{TSEncoder}([z_{\mathrm{CLS}}; \mathbf{E}_{\mathrm{seq}}]),
\end{equation}

\begin{equation}
\mathbf{E}_{\mathrm{Time}}^{\mathrm{Seq}}= {\rm FFN}({\rm Flatten}(\mathbf{E}_{\mathrm{Time}}))
\end{equation}

where $\mathbf{E}_{\mathrm{Time}}^{\mathrm{CLS}}$ denotes the learned CLS representation, and $\mathbf{E}_{\mathrm{Time}}$ contains the contextualized patch representations. The localized features $\mathbf{E}_{\mathrm{Time}}$ are subsequently flattened and projected to form the numerical representation $\mathbf{E}_{\mathrm{Time}}^{\mathrm{Seq}}$, which serves as the query $\psi_{q}$ in the alignment phase.



\subsection{Global-driven Context Learning}
To bridge the gap between the time and frequency domains and fuse multi-modal information, we propose a dynamic contextual mechanism that modulates the prompt space based on the global feature.

We utilize the learned CLS representation $\mathbf{E}_{Time}^{CLS}$ to capture the global characteristics of the input sequence. This global representation is projected through dual transformation heads to generate instance-specific offsets, $\mathcal{C}_K$ and $\mathcal{C}_V$, which act as residual shifts to align the generic soft prompts with the current numerical regime:
\begin{equation}
\mathcal{C}_K = \text{FFN}_{K}(\mathbf{E}_{Time}^{CLS}), \quad \mathcal{C}_V = \text{FFN}_{V}(\mathbf{E}_{Time}^{CLS})
\end{equation}
The calibrated prompts, $\psi_{k}$ and $\psi_{v}$, are then formulated by integrating these offsets into the retrieved soft prompt embeddings $\mathcal{I}_{soft}$. In this way, we can ensure that the semantic priors are conditioned on the specific instance's context:
\begin{equation}
\psi_{k} = \mathcal{I}_{soft} + \mathcal{C}_K, \quad \psi_{v} = \mathcal{I}_{soft} + \mathcal{C}_V
\end{equation}

To fuse representations from different modalities, we employ a Cross-Attention (CA) mechanism. Specifically, the flattened numerical features $\mathbf{E}_{Time}^{Seq}$ serve as the Query, while the calibrated semantic embeddings $\{\psi_{k}, \psi_{v}\}$ function as the Key and Value, respectively. This configuration allows the numerical sequence to selectively attend to the most pertinent semantic prompt. The resulting context-aware representation is integrated with the original temporal features via a residual connection and mapped through a forecasting head to yield the final prediction $\hat{\mathcal{Y}}$:\begin{equation}\hat{\mathcal{Y}} = \text{Head}(\text{CA}(\mathbf{E}_{Time}^{Seq}, \psi_{k}, \psi_{v}) + \mathbf{E}_{Time}^{Seq})\end{equation}




\subsection{Training and Loss function}

Before model training, FreSia first converts the generated frequency-guided prompts into semantic embeddings. Specifically, for each input window, FGPrompt generates a textual prompt that describes the dominant frequency-derived cycles together with auxiliary temporal contexts such as trend and volatility. The prompt is then tokenized and fed into a frozen pretrained language model to obtain its semantic embedding. The language model is not updated during training, and the prompt embeddings can be precomputed and saved before the forecasting model is optimized. Therefore, during training and inference, FreSia only uses the cached semantic embeddings together with the time-series input, avoiding repeated online calls to the language model.

We adopt the Huber loss as the optimization objective to improve robustness against outliers in time-series observations. The Huber loss is defined as:

\begin{equation}
\mathcal{L} = \begin{cases}\frac{1}{2}(\mathcal{Y} - \hat{\mathcal{Y}})^2 & \text{for } |\mathcal{Y} - \hat{\mathcal{Y}}| \le \delta \\ \delta (|\mathcal{Y} - \hat{\mathcal{Y}}| - \frac{1}{2}\delta) & \text{otherwise}\end{cases}
\end{equation}

In our implementation, we set \(\delta=1.0\). Under this setting, the Huber loss is equivalent to the SmoothL1 loss in PyTorch.

\begingroup
\renewcommand{\arraystretch}{0.5}
\begin{table*}[!t]
\centering
\setlength{\tabcolsep}{1mm}{
\begin{tabular}{cc|cc|cc|cc|cc|cc}
\toprule

\multicolumn{2}{c|}{\textbf{Models}}
& \multicolumn{2}{c|}{\textbf{FreSia}} 
& \multicolumn{2}{c|}{LangTime} 
& \multicolumn{2}{c|}{TimeCMA} 
& \multicolumn{2}{c|}{Time-LLM} 
& \multicolumn{2}{c}{UniTime}

\\
\cmidrule(r){1-2}\cmidrule(r){3-4} \cmidrule(r){5-6} \cmidrule(r){7-8}\cmidrule(r){9-10} \cmidrule(r){11-12}
\multicolumn{2}{c}{\textbf{Metrics}} 
& \textbf{MSE} & \textbf{MAE} 
& \textbf{MSE} & \textbf{MAE} 
& \textbf{MSE} & \textbf{MAE}  
& \textbf{MSE} & \textbf{MAE}
& \textbf{MSE} & \textbf{MAE}
\\
\midrule

\multirow{5}{*}{\rotatebox{90}{ETTm1}}
& 96 &\textcolor{red}{0.310} &\textcolor{red}{0.348} &0.319 &\textcolor{red}{0.348} &\textcolor{blue}{0.312} &\textcolor{blue}{0.351} &0.359 &0.381 &0.322 &0.363
\\
& 192 &\textcolor{red}{0.352} &\textcolor{red}{0.374} &0.368 &\textcolor{blue}{0.375} &\textcolor{blue}{0.361} &0.378 &0.383 &0.393 &0.366 &0.387
\\
& 336 &\textcolor{red}{0.378} &\textcolor{red}{0.396} &0.413 &0.402 &\textcolor{blue}{0.392} &\textcolor{blue}{0.401} &0.416 &0.414 &0.398 &0.407
\\
& 720 &\textcolor{red}{0.430} &\textcolor{red}{0.427} &0.487 &0.439 &\textcolor{blue}{0.453} &\textcolor{blue}{0.438} &0.483 &0.449 &0.454 &0.440
\\
&Avg &\textcolor{red}{0.367} &\textcolor{red}{0.386} &0.397 &\textcolor{blue}{0.391} &\textcolor{blue}{0.380} &0.392 &0.410 &0.409 &0.385 &0.399
\\

\midrule
\multirow{5}{*}{\rotatebox{90}{ETTm2}}
& 96  &\textcolor{red}{0.169} &\textcolor{red}{0.251} &0.188 &\textcolor{blue}{0.258} &\textcolor{blue}{0.173} &0.258 &0.193 &0.280 &0.183 &0.266
\\
& 192 &\textcolor{red}{0.237} &\textcolor{red}{0.296} &0.245 &\textcolor{blue}{0.297} &\textcolor{blue}{0.238} &0.301 &0.257 &0.318 &0.251 &0.310
\\
& 336 &\textcolor{blue}{0.298} &\textcolor{red}{0.337} &0.301 &\textcolor{blue}{0.336} &\textcolor{red}{0.297} &0.338 &0.317 &0.353 &0.319 &0.351
\\
& 720 &\textcolor{red}{0.392} &0.395 &\textcolor{blue}{0.402} &\textcolor{red}{0.393} &0.393 &\textcolor{blue}{0.394} &0.419 &0.411 &0.420 &0.410 
\\
&Avg &\textcolor{red}{0.274} &\textcolor{red}{0.320} &0.284 &\textcolor{blue}{0.321} &\textcolor{blue}{0.275} &0.323 &0.296 &0.340 &0.293 &0.334
\\

\midrule
\multirow{5}{*}{\rotatebox{90}{ETTh1}}
& 96  &\textcolor{red}{0.361} &\textcolor{blue}{0.389} &0.391 &\textcolor{red}{0.388} &\textcolor{blue}{0.373} &0.391 &0.398 &0.410 &0.397 &0.418
\\
& 192 &\textcolor{red}{0.406} &\textcolor{red}{0.418} &0.429 &\textcolor{blue}{0.419} &\textcolor{blue}{0.427} &0.421 &0.451 &0.440 &0.434 &0.439 
\\
& 336 &\textcolor{red}{0.419} &\textcolor{red}{0.430} &0.462 &\textcolor{blue}{0.440} &\textcolor{blue}{0.458} &0.448 &0.473 &0.451 &0.468 &0.457
\\
& 720 &\textcolor{red}{0.446} &\textcolor{blue}{0.455} &\textcolor{blue}{0.458} &\textcolor{red}{0.445} &0.449 &0.460 &0.469 &0.470 &0.469 &0.477 
\\
&Avg &\textcolor{red}{0.408} &\textcolor{red}{0.423} &0.435 &\textcolor{blue}{0.423} &\textcolor{blue}{0.423} &0.431 &0.448 &0.443 &0.442 &0.448
\\

\midrule
\multirow{5}{*}{\rotatebox{90}{ETTh2}}
& 96  &\textcolor{red}{0.270} &\textcolor{red}{0.328} &0.299 &\textcolor{blue}{0.336} &\textcolor{blue}{0.286} &\textcolor{blue}{0.336} &0.295 &0.345 &0.296 &0.345 
\\
& 192 &\textcolor{red}{0.347} &\textcolor{red}{0.380} &0.374 &\textcolor{blue}{0.382} &\textcolor{blue}{0.363} &0.387 &0.386 &0.399 &0.374 &0.394
\\
& 336 &\textcolor{red}{0.355} &\textcolor{red}{0.394} &0.410 &\textcolor{blue}{0.418} &\textcolor{blue}{0.406} &0.421 &0.419 &0.429 &0.415 &0.427
\\
& 720 &\textcolor{red}{0.400} &\textcolor{blue}{0.428} &0.418 &\textcolor{red}{0.426} &\textcolor{blue}{0.417} &0.438 &0.425 &0.442 &0.425 &0.444 
\\
&Avg &\textcolor{red}{0.343} &\textcolor{red}{0.382} &0.375 &\textcolor{blue}{0.391} &\textcolor{blue}{0.372} &0.397 &0.381 &0.404 &0.378 &0.403
\\

\midrule
\multirow{5}{*}{\rotatebox{90}{Electricity}}
& 96 &\textcolor{red}{0.145} &\textcolor{red}{0.243} &0.181 &0.266 &\textcolor{blue}{0.146} &\textcolor{blue}{0.245} &0.172 &0.265 &0.196 &0.287
\\
& 192 &\textcolor{blue}{0.163} &\textcolor{blue}{0.260} &0.185 &0.273 &\textcolor{blue}{0.165} &\textcolor{blue}{0.264} &0.182 &0.279 &0.199 &0.291
\\
& 336 &\textcolor{red}{0.180} &\textcolor{red}{0.276} &0.198 &\textcolor{blue}{0.281} &0.235 &0.336 &\textcolor{blue}{0.195} &0.288 &0.214 &0.305
\\
& 720 &\textcolor{red}{0.212} &\textcolor{red}{0.306} &0.241 &0.320 &\textcolor{blue}{0.322} &\textcolor{blue}{0.403} &0.233 &0.320 &0.254 &0.335
\\
&Avg &\textcolor{red}{0.175} &\textcolor{red}{0.271} &0.201 &\textcolor{blue}{0.285} &0.217 &0.312 &\textcolor{blue}{0.195} &0.288 &0.216 &0.306
\\

\midrule
\multirow{5}{*}{\rotatebox{90}{Weather}}
& 96  &\textcolor{red}{0.150} &\textcolor{red}{0.192} &0.178 &\textcolor{blue}{0.202} &\textcolor{blue}{0.167} &0.211 &0.198 &0.235 &0.171 &0.214
\\
& 192 &\textcolor{red}{0.202} &\textcolor{red}{0.241} &\textcolor{blue}{0.211} &\textcolor{blue}{0.245} &0.212 &0.253 &0.240 &0.269 &0.217 &0.254 
\\
& 336 &\textcolor{red}{0.259} &\textcolor{red}{0.284} &\textcolor{blue}{0.269} &\textcolor{blue}{0.286} &0.270 &0.292 &0.295 &0.308 &0.274 &0.293
\\
& 720 &\textcolor{red}{0.336} &\textcolor{red}{0.335} &0.351 &\textcolor{blue}{0.346} &\textcolor{blue}{0.350} &0.348 &0.368 &0.353 &0.351 &0.343
\\
&Avg &\textcolor{red}{0.236} &\textcolor{red}{0.263} &0.252 &\textcolor{blue}{0.270} &\textcolor{blue}{0.250} &0.276 &0.275 &0.291 &0.253 &0.276 
\\

\midrule
\multirow{5}{*}{\rotatebox{90}{Exchange}}
& 96 &\textcolor{red}{0.081} &\textcolor{red}{0.200} &0.089 &\textcolor{blue}{0.201} &0.115 &0.243 &0.087 &0.208 &0.086 &0.209
\\
& 192 &0.177 &\textcolor{red}{0.299} &\textcolor{blue}{0.175} &\textcolor{blue}{0.298} &0.209 &0.330 &0.178 &0.302 &\textcolor{red}{0.174} &0.299
\\
& 336 &\textcolor{red}{0.318} &\textcolor{red}{0.407} &0.329 &0.409 &0.374 &0.449 &0.338 &0.422 &\textcolor{blue}{0.319} &\textcolor{blue}{0.408}
\\
& 720 &\textcolor{red}{0.781} &\textcolor{red}{0.668} &0.852 &\textcolor{blue}{0.690} &1.083 &0.803 &\textcolor{blue}{0.819} &\textcolor{blue}{0.681} &0.875 &0.701 
\\
&Avg &\textcolor{red}{0.339} &\textcolor{red}{0.394} &0.361 &\textcolor{blue}{0.400} &0.445 &0.456 &\textcolor{blue}{0.356} &0.403 &0.364 &0.404
\\

\midrule
\multirow{5}{*}{\rotatebox{90}{ILI}}
&24  &\textcolor{red}{1.580} &\textcolor{red}{0.792} &4.295 &1.537 &\textcolor{blue}{1.996} &\textcolor{blue}{0.998} &2.383 &1.004 &2.346 &0.954
\\
&36 &\textcolor{red}{1.373} &\textcolor{red}{0.717} &4.272 &1.588 &\textcolor{blue}{1.906} &\textcolor{blue}{0.915} &2.390 &0.993 &1.998 &0.912 
\\
&48 &\textcolor{red}{1.730} &\textcolor{red}{0.803} &4.264 &1.552 &\textcolor{blue}{1.867} &\textcolor{blue}{0.868} &2.394 &1.003 &1.979 &0.912
\\
&60 &\textcolor{red}{1.568} &\textcolor{red}{0.802} &4.076 &1.518 &\textcolor{blue}{1.920} &\textcolor{blue}{0.904} &2.562 &1.049 &2.109 &0.938
\\
&Avg &\textcolor{red}{1.563} &\textcolor{red}{0.779} &4.227 &1.549 &\textcolor{blue}{1.922} &\textcolor{blue}{0.921} &2.432 &1.012 &2.108 &0.929 
\\

\bottomrule
\end{tabular}
}

\caption{Results of multivariate time series forecasting. The input sequence length is 36 for the ILI dataset and 96 for others. The best results are in red and the second best are blue.}
\label{table1}
\end{table*}
\endgroup

\begingroup
\renewcommand{\arraystretch}{0.5}
\begin{table*}[!t]
\centering
\setlength{\tabcolsep}{1mm}{
\begin{tabular}{cc|cc|cc|cc|cc|cc}
\toprule

\multicolumn{2}{c|}{\textbf{Models}}
& \multicolumn{2}{c|}{\textbf{FreSia}} 
& \multicolumn{2}{c|}{OFA}
& \multicolumn{2}{c|}{iTransformer} 
& \multicolumn{2}{c|}{TimesNet}
& \multicolumn{2}{c}{RLinear}

\\
\cmidrule(r){1-2}\cmidrule(r){3-4} \cmidrule(r){5-6} \cmidrule(r){7-8}\cmidrule(r){9-10} \cmidrule(r){11-12} 
\multicolumn{2}{c|}{\textbf{Metrics}} 
& \textbf{MSE} & \textbf{MAE} 
& \textbf{MSE} & \textbf{MAE} 
& \textbf{MSE} & \textbf{MAE}  
& \textbf{MSE} & \textbf{MAE}
& \textbf{MSE} & \textbf{MAE}
\\
\midrule

\multirow{5}{*}{\rotatebox{90}{ETTm1}}
& 96 &\textcolor{red}{0.310} &\textcolor{red}{0.348} &0.335 &0.369 &\textcolor{blue}{0.334} &\textcolor{blue}{0.368} &0.338 &0.375 &0.355 &0.376
\\
& 192 &\textcolor{red}{0.352} &\textcolor{red}{0.374} &\textcolor{blue}{0.374} &\textcolor{blue}{0.385} &0.377 &0.391 &0.374 &0.387 &0.387 &0.392
\\
& 336 &\textcolor{red}{0.378} &\textcolor{red}{0.396} &\textcolor{blue}{0.407} &\textcolor{blue}{0.406} &0.426 &0.420 &0.410 &0.411 &0.424 &0.415
\\
& 720 &\textcolor{red}{0.430} &\textcolor{red}{0.427} &\textcolor{blue}{0.469} &\textcolor{blue}{0.442} &0.491 &0.459 &0.478 &0.450 &0.487 &0.450
\\
&Avg &\textcolor{red}{0.367} &\textcolor{red}{0.386} &\textcolor{blue}{0.396} &\textcolor{blue}{0.401} &0.407 &0.410 &0.400 &0.406 &0.413 &0.408
\\

\midrule
\multirow{5}{*}{\rotatebox{90}{ETTm2}}
& 96  &\textcolor{red}{0.169} &\textcolor{red}{0.251} &0.190 &0.275 &\textcolor{blue}{0.180} &\textcolor{blue}{0.264} &0.187 &0.267 &0.182 &0.265
\\
& 192 &\textcolor{red}{0.237} &\textcolor{red}{0.296} &0.253 &0.313 &0.250 &0.309 &0.249 &0.309 &\textcolor{blue}{0.246} &\textcolor{blue}{0.304}
\\
& 336 &\textcolor{red}{0.298} &\textcolor{red}{0.337} &0.321 &0.360 &0.311 &0.348 &0.321 &0.351 &\textcolor{blue}{0.307} &\textcolor{blue}{0.342}
\\
& 720 &\textcolor{red}{0.392} &\textcolor{red}{0.395} &\textcolor{blue}{0.402} &0.406 &0.412 &0.407 &0.408 &0.403 &0.407 &\textcolor{blue}{0.398}
\\
&Avg &\textcolor{red}{0.274} &\textcolor{red}{0.320} &0.294 &0.339 &0.288 &0.332 &0.291 &0.333 &\textcolor{blue}{0.286} &\textcolor{blue}{0.327}
\\

\midrule
\multirow{5}{*}{\rotatebox{90}{ETTh1}}
& 96  &\textcolor{red}{0.361} &\textcolor{red}{0.389} &0.398 &0.424 &0.386 &0.405 &\textcolor{blue}{0.384} &0.402 &0.386 &\textcolor{blue}{0.395}
\\
& 192 &\textcolor{red}{0.406} &\textcolor{red}{0.418} &0.449 &0.427 &0.441 &0.436 &\textcolor{blue}{0.434} &0.429 &0.437 &\textcolor{blue}{0.424}
\\
& 336 &\textcolor{red}{0.419} &\textcolor{red}{0.430} &0.492 &0.466 &0.487 &0.458 &0.491 &0.469 &\textcolor{blue}{0.479} &\textcolor{blue}{0.446}
\\
& 720 &\textcolor{red}{0.446} &\textcolor{red}{0.455} &0.487 &0.483 &0.503 &0.491 &0.521 &0.500 &\textcolor{blue}{0.481} &\textcolor{blue}{0.470}
\\
&Avg &\textcolor{red}{0.408} &\textcolor{red}{0.423} &0.457 &0.450 &0.454 &0.447 &0.458 &0.450 &\textcolor{blue}{0.446} &\textcolor{blue}{0.434}
\\

\midrule
\multirow{5}{*}{\rotatebox{90}{ETTh2}}
& 96  &\textcolor{red}{0.270} &\textcolor{red}{0.328} &0.312 &0.360 &\textcolor{blue}{0.297} &\textcolor{blue}{0.349} &0.340 &0.374 &0.318 &0.363
\\
& 192 &\textcolor{red}{0.347} &\textcolor{red}{0.380} &0.387 &0.405 &\textcolor{blue}{0.380} &\textcolor{blue}{0.400} &0.402 &0.414 &0.401 &0.412
\\
& 336 &\textcolor{red}{0.355} &\textcolor{red}{0.394} &\textcolor{blue}{0.424} &0.437 &0.428 &\textcolor{blue}{0.432} &0.452 &0.452 &0.436 &0.442
\\
& 720 &\textcolor{red}{0.400} &\textcolor{blue}{0.428} &0.433 &0.453 &\textcolor{blue}{0.427} &\textcolor{blue}{0.445} &0.462 &0.468 &0.442 &0.454
\\
&Avg &\textcolor{red}{0.343} &\textcolor{red}{0.382} &0.389 &0.414 &\textcolor{blue}{0.383} &\textcolor{blue}{0.407} &0.414 &0.427 &0.399 &0.418
\\

\midrule
\multirow{5}{*}{\rotatebox{90}{Electricity}}
& 96 &\textcolor{red}{0.145} &\textcolor{blue}{0.243} &0.197 &0.290 &\textcolor{blue}{0.148} &\textcolor{red}{0.240} &0.168 &0.272 &0.201 &0.281
\\
& 192 &\textcolor{blue}{0.163} &\textcolor{blue}{0.260} &0.201 &0.292 &\textcolor{red}{0.162} &\textcolor{red}{0.253} &0.184 &0.289 &0.201 &0.283
\\
& 336 &\textcolor{blue}{0.180} &\textcolor{blue}{0.276} &0.217 &0.309 &\textcolor{red}{0.178} &\textcolor{red}{0.269} &0.198 &0.300 &0.215 &0.298
\\
& 720 &\textcolor{red}{0.212} &\textcolor{red}{0.306} &0.253 &0.339 &0.225 &\textcolor{blue}{0.317} &\textcolor{blue}{0.220} &0.320 &0.257 &0.331
\\
&Avg &\textcolor{red}{0.175} &\textcolor{blue}{0.271} &0.217 &0.308 &\textcolor{blue}{0.178} &\textcolor{red}{0.270} &0.192 &0.295 &0.219 &0.298
\\

\midrule
\multirow{5}{*}{\rotatebox{90}{Weather}}
& 96  &\textcolor{red}{0.150} &\textcolor{red}{0.192} &0.203 &0.244 &0.174 &\textcolor{blue}{0.214} &\textcolor{blue}{0.172} &0.220 &0.192 &0.232
\\
& 192 &\textcolor{red}{0.202} &\textcolor{red}{0.241} &0.247 &0.277 &0.221 &\textcolor{blue}{0.254} &\textcolor{blue}{0.219} &0.261 &0.240 &0.271
\\
& 336 &\textcolor{red}{0.259} &\textcolor{red}{0.284} &0.297 &0.311 &\textcolor{blue}{0.278} &\textcolor{blue}{0.296} &0.280 &0.306 &0.292 &0.307
\\
& 720 &\textcolor{red}{0.336} &\textcolor{red}{0.335} &0.368 &0.356 &\textcolor{blue}{0.358} &\textcolor{blue}{0.349} &0.365 &0.359 &0.364 &0.353
\\
&Avg &\textcolor{red}{0.236} &\textcolor{red}{0.263} &0.279 &0.297 &\textcolor{blue}{0.258} &\textcolor{blue}{0.278} &0.259 &0.287 &0.272 &0.291
\\

\midrule
\multirow{5}{*}{\rotatebox{90}{Exchange}}
& 96 &\textcolor{red}{0.081} &\textcolor{red}{0.200} &\textcolor{blue}{0.084} &\textcolor{blue}{0.202} &0.086 &0.206 &0.107 &0.234 &0.093 &0.217
\\
& 192 &\textcolor{red}{0.177} &\textcolor{blue}{0.299} &\textcolor{red}{0.177} &\textcolor{red}{0.298} &\textcolor{red}{0.177} &0.299 &0.226 &0.344 &\textcolor{blue}{0.184} &0.307
\\
& 336 &\textcolor{red}{0.318} &\textcolor{red}{0.407} &0.339 &0.420 &\textcolor{blue}{0.331} &\textcolor{blue}{0.417} &0.367 &0.448 &0.351 &0.432
\\
& 720 &\textcolor{red}{0.781} &\textcolor{red}{0.668} &0.856 &0.696 &\textcolor{blue}{0.847} &\textcolor{blue}{0.691} &0.964 &0.746 &0.886 &0.714
\\
&Avg &\textcolor{red}{0.339} &\textcolor{red}{0.394} &0.364 &0.404 &\textcolor{blue}{0.360} &\textcolor{blue}{0.403} &0.416 &0.443 &0.379 &0.418
\\

\midrule
\multirow{5}{*}{\rotatebox{90}{ILI}}
&24  &\textcolor{red}{1.580} &\textcolor{red}{0.792} &2.732 &1.100 &2.347 &1.731 &\textcolor{blue}{2.317} &\textcolor{blue}{0.934} &2.436 &0.970
\\
&36 &\textcolor{red}{1.373} &\textcolor{red}{0.717} &2.664 &1.063 &2.468 &0.998 &\textcolor{blue}{1.972} &\textcolor{blue}{0.920} &2.362 &0.953
\\
&48 &\textcolor{red}{1.730} &\textcolor{red}{0.803} &2.617 &1.041 &2.489 &1.016 &\textcolor{blue}{2.238} &\textcolor{blue}{0.913} &2.461 &0.982
\\
&60 &\textcolor{red}{1.568} &\textcolor{red}{0.802} &2.478 &1.035 &2.471 &1.065 &\textcolor{blue}{2.027} &\textcolor{blue}{0.928} &2.451 &0.979
\\
&Avg &\textcolor{red}{1.563} &\textcolor{red}{0.779} &2.623 &1.060 &2.444 &1.203 &\textcolor{blue}{2.139} &\textcolor{blue}{0.924} &2.428 &0.971
\\

\bottomrule
\end{tabular}
}

\caption{Results of multivariate time series forecasting. The input sequence length is 36 for the ILI dataset and 96 for others. The best results are in red and the second best are blue.}
\label{table11}
\end{table*}
\endgroup

\section{Experiments}
\subsection{Experimental Setup}
\paragraph{Datasets and Evaluation Metrics} For MTSF, we evaluate FreSia on 8 datasets: four ETT (ETTh1/h2, ETTm1/m2)~\cite{zhou2021informer}, Exchange~\cite{lai2018modeling}, Weather~\cite{wu2021autoformer}, Electricity~\cite{wu2021autoformer}, and the ILI~\cite{wu2021autoformer} dataset. We use Mean Squared Error (MSE) and Mean Absolute Error (MAE) as evaluation metrics. For anomaly detection, we evaluate on five datasets: SMD~\cite{su2019robust}, PSM~\cite{abdulaal2021practical}, SWaT~\cite{mathur2016swat}, MSL~\cite{hundman2018detecting}, SMAP~\cite{hundman2018detecting}, using precision (P), recall (R), and F1-score (F1) as evaluation metrics.

\begingroup
\renewcommand{\arraystretch}{0.5}
\begin{table*}[!tt]
\centering
\setlength{\tabcolsep}{1mm}{
\begin{tabular}{c|c|c|c|c|c|c|c|c}
\toprule

\multicolumn{1}{c|}{\textbf{Dataset}}
&\multicolumn{1}{c|}{\textbf{Metric}}
& \multicolumn{1}{c|}{\textbf{FreSia}} 
& \multicolumn{1}{c|}{TSINR} 
& \multicolumn{1}{c|}{FPT} 
& \multicolumn{1}{c|}{TimesNet} 
& \multicolumn{1}{c|}{Anomaly.}
& \multicolumn{1}{c|}{LightTS}
& \multicolumn{1}{c}{DLinear}

\\
\midrule

\multirow{3}{*}{SMD}
& P &\textcolor{red}{88.56} &83.09 &87.27 &\textcolor{blue}{87.98} &78.72 &87.04 &87.27
\\
& R &\textcolor{red}{84.07} &80.46 &81.08 &\textcolor{blue}{81.54} &65.43 &78.39 &80.99
\\
& F1 &\textcolor{red}{86.25} &81.76 &84.06 &\textcolor{blue}{84.64} &71.46 &82.49 &84.01
\\

\midrule
\multirow{3}{*}{PSM}
& P &97.61 &\textcolor{red}{99.21} &98.55 &98.51 &\textcolor{blue}{98.76} &98.29  &98.66
\\
& R &\textcolor{blue}{96.22} &89.37 &95.79 &\textcolor{red}{96.27} &83.25 &93.60 &94.70
\\
& F1 &96.91 &94.04 &\textcolor{blue}{97.15} &\textcolor{red}{97.38} &90.35 &95.89 &96.64
\\

\midrule
\multirow{3}{*}{SWaT}
& P &96.08 &\textcolor{blue}{99.31} &92.12 &90.75 &\textcolor{red}{99.73} &92.36 &80.91
\\
& R &92.64 &72.32 &93.06 &\textcolor{red}{95.40} &68.07 &93.32 &\textcolor{blue}{95.30}
\\
&F1 &\textcolor{red}{94.33} &83.69 &92.59 &93.02 & 80.91 &\textcolor{blue}{92.84} &87.52
\\

\midrule
\multirow{3}{*}{MSL}
& P &87.19 &83.57 &82.03 &89.55 &89.78 &\textcolor{blue}{89.17} &\textcolor{red}{89.68}  
\\
& R &\textcolor{blue}{84.09} &\textcolor{red}{85.40} &82.01 &75.29 &73.66 &73.64 &75.31
\\
& F1 &\textcolor{red}{85.61} &\textcolor{blue}{84.47} &82.02 &81.80 &80.93 &80.66 &81.87
\\

\midrule
\multirow{3}{*}{SMAP}
& P &88.47 &87.55 &\textcolor{red}{90.91} &89.92 &\textcolor{blue}{90.14} &90.02 &89.89
\\
& R &\textcolor{red}{70.90} &55.32 &\textcolor{blue}{61.01} &56.56 &54.00 &53.90 &54.01
\\
& F1 &\textcolor{red}{78.72} &67.80 &\textcolor{blue}{73.02} &69.44 &67.54 &67.43 &67.48
\\

\midrule
\multirow{3}{*}{Avg}
& P &\textcolor{red}{91.58} &90.55 &90.18 &91.34 &80.25 &91.38 &\textcolor{blue}{91.55}
\\
& R &\textcolor{red}{85.58} &76.57 &82.59 &81.01 &\textcolor{blue}{83.95} &78.57 &79.62
\\
& F1 &\textcolor{red}{88.36} &82.35 &\textcolor{blue}{85.77} &85.26 &80.50 &83.86 &84.54
\\

\bottomrule
\end{tabular}
}

\caption{Results of anomaly detection. The input sequence length is 100. The best results are in red and the second best are blue.}
\label{tab:anomaly_detection}
\end{table*}
\endgroup

\paragraph{Baselines} For MTSF, we select eight representative forecasting models as baselines. These include (1) Prompt-based LLMs: LangTime~\cite{pmlr-v267-niu25e}, TimeCMA~\cite{liu2025timecma}, Time-LLM~\cite{ICLR2024_680b2a81}, UniTime~\cite{liu2024unitime}; (2) LLM-based: OFA~\cite{zhou2023one}; (3) Transformer-based model: iTransformer~\cite{liu2024iclr-itransformer}; (4) CNN-based method: TimesNet~\cite{DBLP:conf/iclr/WuHLZ0L23}; (5) Linear-based method: RLinear~\cite{DBLP:journals/corr/abs-2305-10721}. For anomaly detection, we use two kinds of models, including: (1) general frameworks designed for time series modeling: FPT~\cite{zhou2023one}, TimesNet~\cite{DBLP:conf/iclr/WuHLZ0L23}, LightTS~\cite{DBLP:journals/corr/abs-2207-01186} and DLinear~\cite{zeng2023transformers}; (2) models specifically tailored for time series anomaly detection: TSINR~\cite{li2025tsinr}, AnomalyTransformer~\cite{xu2022iclr-anomaly}.
All experiments are implemented in PyTorch 2.1.2 and on an NVIDIA RTX 4090.


\subsection{Main Results}
\paragraph{Long-term Forecasting}
Table \ref{table1} and Table \ref{table11} report the forecasting results on eight datasets. At the dataset-average level, FreSia achieves the lowest MSE on all eight datasets and the lowest or tied-lowest MAE on seven of them. In particular, the improvement is more pronounced on ILI, where FreSia obtains an average MSE/MAE of 1.563/0.779, compared with 1.922/0.921 for the strongest competing result, corresponding to reductions of approximately 18.7\% and 15.4\%, respectively. FreSia also achieves average MSE/MAE values of 0.343/0.382 on ETTh2, 0.236/0.263 on Weather, and 0.339/0.394 on Exchange, achieving lower average errors than the compared LLM-based and conventional forecasting models on these datasets. On Electricity, FreSia obtains the lowest average MSE of 0.175, while its MAE of 0.271 is marginally higher than the 0.270 achieved by iTransformer. Overall, these results indicate that the performance gains are observed across datasets with different sampling frequencies and temporal characteristics, rather than being concentrated on a single benchmark. We further compare FreSia with representative frequency-domain baselines, including TimeKAN, PaiFilter, TexFilter, and FredFormer. The detailed results are reported in~\ref{app:frequency baseline}.

\paragraph{Anomaly Detection} 
Table~\ref{tab:anomaly_detection} reports the anomaly detection results on five multivariate benchmarks. FreSia achieves the best average performance with 91.58\% precision, 85.58\% recall, and 88.36\% F1 score, outperforming the strongest competing baseline FPT by 2.59 percentage points in average F1. At the dataset level, FreSia obtains the best F1 score on four out of five datasets, including SMD, SWaT, MSL, and SMAP. Notably, FreSia achieves the highest average recall among all methods, indicating that it can better identify anomalous patterns while maintaining competitive precision. These results suggest that the learned frequency-semantic representations are not limited to forecasting, but also generalize well to reconstruction-based anomaly detection. We further provide a diagnostic comparison of different scoring rules in~\ref{app:anomaly_score}.

\subsection{Sensitivity of Frequency Pool Size}
\begin{figure*}[h]
\centering
\includegraphics[width=0.5\textwidth]{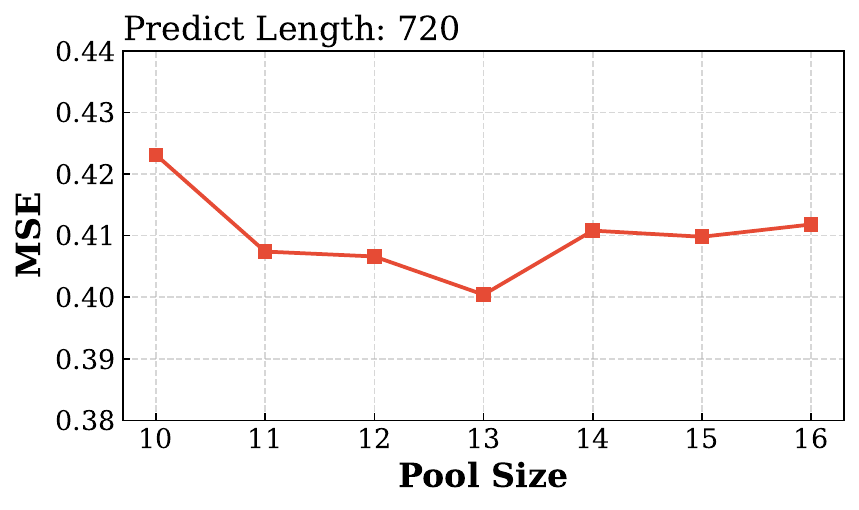}
\caption{Sensitivity analysis of frequency pool size.}
\label{fig9}
\end{figure*}

We conduct parameter sensitivity analysis on the frequency pool size with ETTh2 720 horizon. As can be seen from Figure~\ref{fig9}, the frequency pool size affects the prediction results. If the frequency pool size is too small, it may not fully represent the latent features in the time series. If it is too large, it will cause redundancy and also affect the prediction results.

\subsection{Ablation}
Table \ref{table3} presents the ablation study results. We evaluate the contribution of each module by averaging metrics across all forecasting horizons.


\paragraph{Core Backbones} Both the LLM and Time-Series Encoder (TSE) contribute to forecasting performance. Removing the LLM (w/o LLM) increases the prediction errors, indicating the contribution of its semantic representations. Removing the TSE (w/o TSE) results in the largest degradation among the evaluated variants, highlighting the importance of temporal dependency modeling.

\begin{table}[h]
\renewcommand{\arraystretch}{0.85}
\centering
\begin{tabular}{c|cc|cc|cc}
\toprule
\textbf{Datasets}
& \multicolumn{2}{c|}{ETTh2} 
& \multicolumn{2}{c|}{ETTm2} 
& \multicolumn{2}{c}{ECL} 
\\
\cmidrule(r){1-1} \cmidrule(r){2-3} \cmidrule(r){4-5} \cmidrule(r){6-7}
\textbf{Metrics} 
& \textbf{MSE} & \textbf{MAE} 
& \textbf{MSE} & \textbf{MAE} 
& \textbf{MSE} & \textbf{MAE} 
\\

\midrule
w/o LLM &0.357 &0.388 &0.288 &0.331 &0.186 &0.271 
\\

w/o TSE &0.393 &0.417 &0.321 &0.358 &0.382 &0.424
\\

w/o PE &0.353 &0.387 &0.279 &0.325 &0.177 &0.274
\\

w/o CA &0.356 &0.390 &0.291 &0.333 &0.179 &0.268
\\

w/o CLS &0.356 &0.388 &0.282 &0.327 &0.179 &0.276
\\

w/o Pool &0.352 &0.387 &0.279 &0.324 &0.183 &0.278
\\

\textbf{FreSia} &\textbf{0.343} &\textbf{0.382} &\textbf{0.274} &\textbf{0.320} &\textbf{0.175} &\textbf{0.271}
\\

\bottomrule

\end{tabular}
\caption{Ablation studies of FreSia}
\label{table3}
\end{table}

\paragraph{Prompting and Pooling} The Prompt Encoder (w/o PE) has a relatively minor impact, yet it facilitates cross-modal alignment. In contrast, excluding the Frequency Pool (w/o Pool) consistently impairs forecasting precision. This suggests that the frequency-domain prototype pool contributes to effective retrieval.

\paragraph{Alignment Mechanisms} The alignment components provide consistent gains. Removing CLS-driven calibration (w/o CLS) reduces accuracy on several datasets, showing that global profiling helps handle regime shifts. Additionally, replacing Cross-Attention (w/o CA) yields inferior results, suggesting that attention-based fusion is more effective than simple concatenation for modeling complex interactions.

\subsection{Analysis}

\begingroup
\renewcommand{\arraystretch}{0.85}
\begin{table}[!t]
\centering
\small
\setlength{\tabcolsep}{3.5pt}{
\begin{tabular}{c|cc|cc|cc|cc}
\toprule
\textbf{Dataset}
& \multicolumn{2}{c|}{\textbf{FreSia}}
& \multicolumn{2}{c|}{TimeCMA-style}
& \multicolumn{2}{c|}{w/o Freq. Path.}
& \multicolumn{2}{c}{w/o LLM}
\\
\cmidrule(lr){2-3}
\cmidrule(lr){4-5}
\cmidrule(lr){6-7}
\cmidrule(lr){8-9}
& \textbf{MSE} & \textbf{MAE}
& \textbf{MSE} & \textbf{MAE}
& \textbf{MSE} & \textbf{MAE}
& \textbf{MSE} & \textbf{MAE}
\\
\midrule
Weather
& \textbf{0.236} & \textbf{0.263}
& 0.237 & 0.264
& 0.238 & 0.264
& 0.252 & 0.273
\\
Exchange
& \textbf{0.339} & \textbf{0.394}
& 0.383 & 0.417
& 0.392 & 0.420
& 0.404 & 0.425
\\
ILI
& \textbf{1.563} & 0.779
& 1.577 & \textbf{0.775}
& 1.601 & 0.782
& 1.696 & 0.787
\\
\midrule
Avg.
& \textbf{0.713} & \textbf{0.479}
& 0.732 & 0.485
& 0.743 & 0.489
& 0.784 & 0.495
\\
\bottomrule
\end{tabular}
}
\caption{Disentangling the effects of frequency-domain information and LLM semantic encoding.}
\label{tab:disentangle_freq_llm}
\end{table}
\endgroup

\paragraph{Disentangling Frequency Information and Semantic Encoding}
Table~\ref{tab:disentangle_freq_llm} reports results on representative datasets with different temporal characteristics, including Weather, Exchange, and ILI, for a compact diagnostic analysis. Removing the LLM encoder increases the average error from 0.713/0.479 to 0.784/0.495, indicating that the frozen LLM provides useful semantic representations. Meanwhile, the w/o Freq. Path. variant retains the semantic encoder but removes frequency descriptors in prompts and disables the frequency-related pool, which degrades the average performance to 0.743/0.489. This shows that the gains do not come solely from the LLM encoder; explicit frequency-domain information also contributes to performance.
We further compare with a TimeCMA-style prompt, which replaces the frequency-guided prompt with a raw-value-style textual prompt. Its average performance, 0.732/0.485, is worse than FreSia, suggesting that frequency-semantic instantiation provides more effective structural cues than raw numerical prompt serialization. The degradation is more pronounced on Exchange and ILI, while Weather shows a milder difference. Overall, these results demonstrate that FreSia benefits from the combination of frequency-domain structural cues and LLM-based semantic encoding.


\begin{figure*}[h]
\centering
\includegraphics[width=0.5\textwidth]{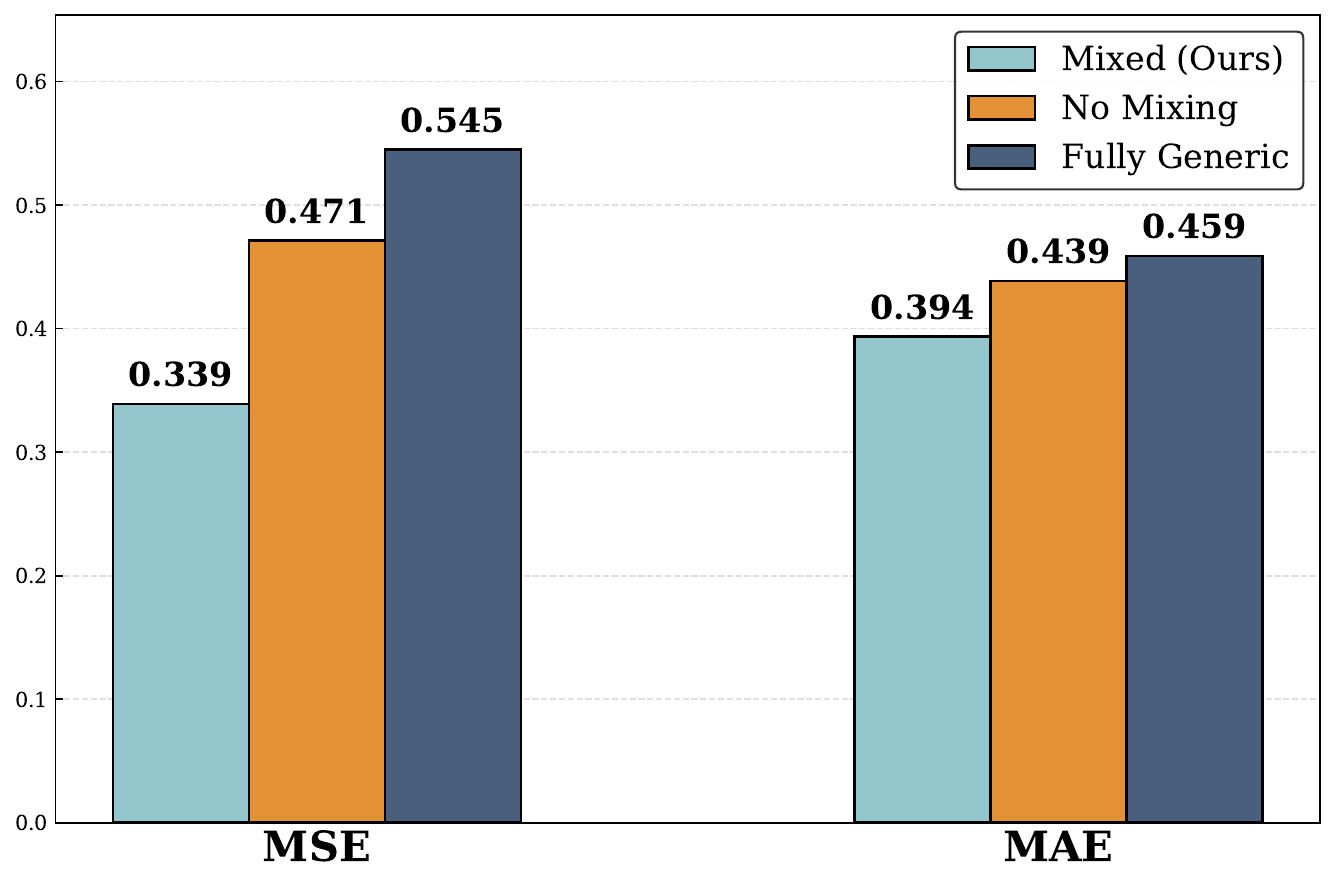}
\caption{Effect of mixed prompting strategy on Exchange.}
\label{fig3}
\end{figure*}

\paragraph{General Prompting Strategies} 

We further analyze how general prompts interact with task-specific frequency prompts. As shown in Figure~\ref{fig3}, the mixed prompting strategy achieves the lowest errors on the Exchange dataset, with 0.339 MSE and 0.394 MAE. Compared with the no-mixing setting, it reduces MSE and MAE by 28.0\% and 10.3\%, respectively. Compared with the fully generic prompting setting, it reduces MSE and MAE by 37.8\% and 14.2\%, respectively.

These results indicate that general prompts are beneficial only when combined with task-specific frequency information. Using general prompts alone introduces overly generic semantic cues, while removing them entirely limits the contextual richness of the prompt. In contrast, the mixed prompting strategy provides both frequency-aware temporal structure and general semantic context, leading to more accurate predictions.

\begin{figure*}[!h]
\centering
\begin{subfigure}{0.32\textwidth}
    \centering
    \includegraphics[width=1\textwidth]{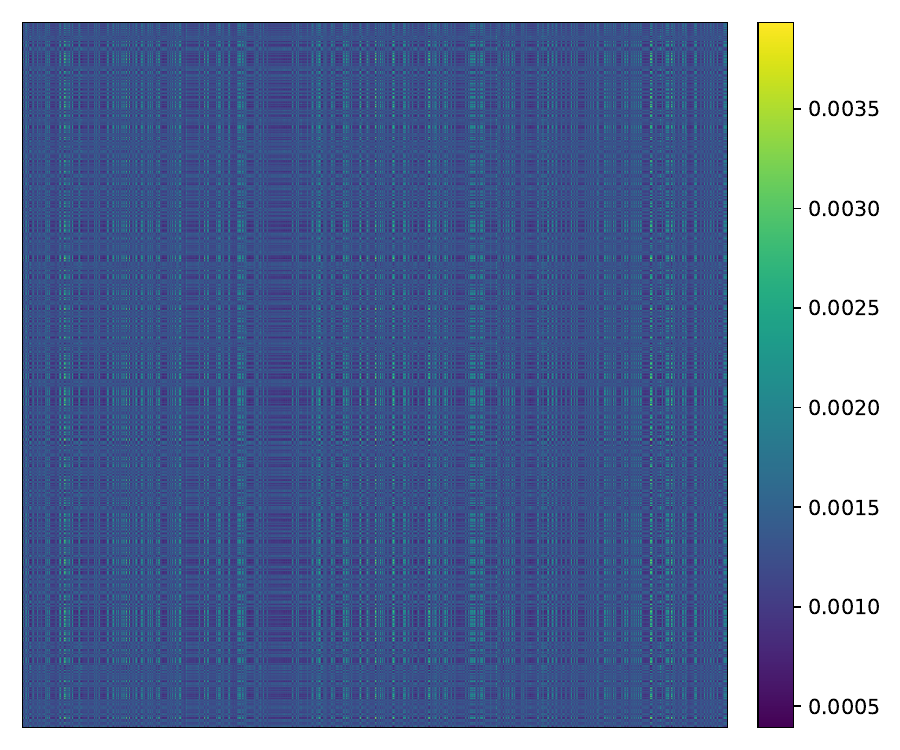}
    \subcaption{Attention With CLS}
\end{subfigure}
\begin{subfigure}{0.32\textwidth}
    \centering
    \includegraphics[width=1\textwidth]{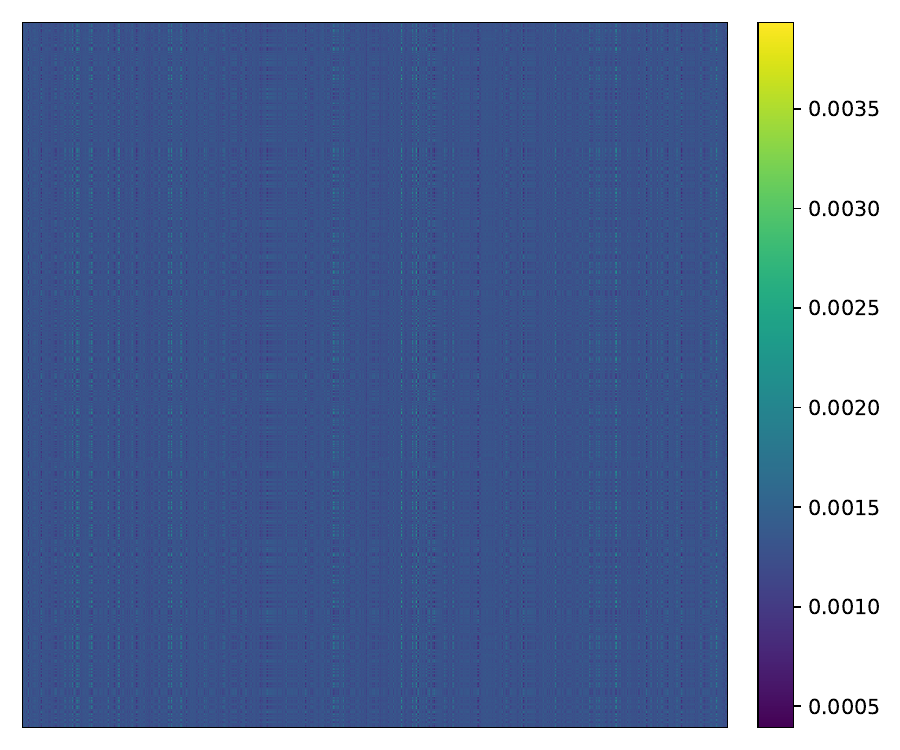}
    \subcaption{Attention without CLS}
\end{subfigure}
\begin{subfigure}{0.33\textwidth}
    \centering
    \includegraphics[width=1\textwidth]{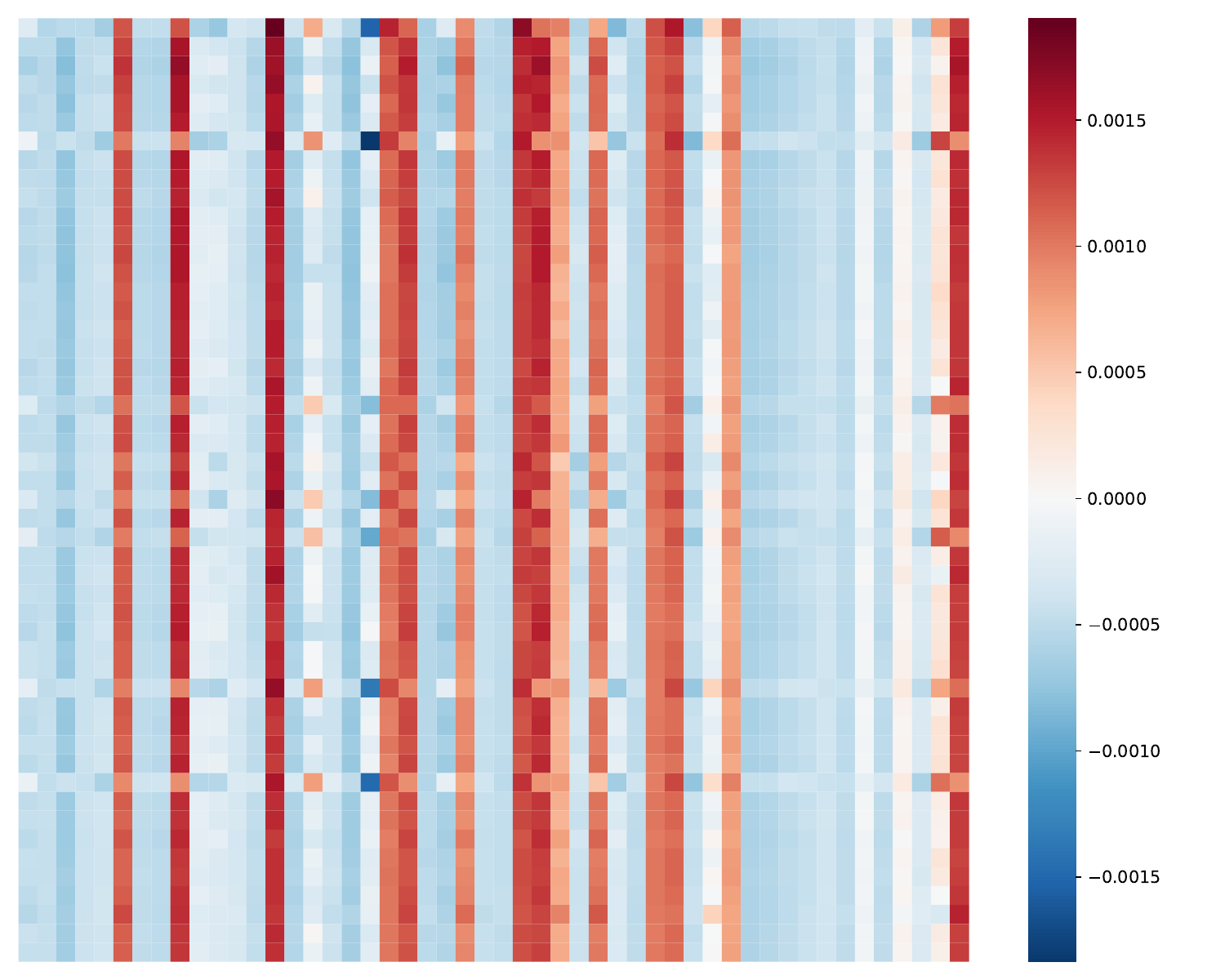}
    \subcaption{Topk difference in attention}
\end{subfigure}
\caption{Attention difference analysis of CLS-driven calibration.}
\label{fig5}
\vspace{-2mm}
\end{figure*}

\paragraph{Mechanistic Analysis of CLS} To evaluate the impact of the CLS token, we analyze the attention within the Cross-Attention (CA) mechanism. Specifically, we visualize the residual attention map (the difference between configurations with and without CLS) and focus on the 50 channels exhibiting the most pronounced variations.


As illustrated in Figure~\ref{fig5} (a) and (b), the attention differences are not uniformly distributed across channels. While several channels exhibit noticeable changes after introducing the CLS-driven calibration, others remain relatively stable. This suggests that the CLS token may help the model adaptively reweight channel-level responses rather than perturbing all representations equally. In Figure~\ref{fig5} (c), the observed local patterns of positive and negative attention changes indicate that CLS-driven calibration can modulate the interaction between semantic prompts and temporal features in a structured manner. These observations provide qualitative evidence that the CLS token contributes to context-aware alignment between global temporal information and semantic prompts, which is consistent with the performance degradation observed when removing CLS in the ablation study.


\begin{figure*}[h]
\centering
\begin{subfigure}{0.33\textwidth}
    \centering
    \includegraphics[width=1\textwidth]{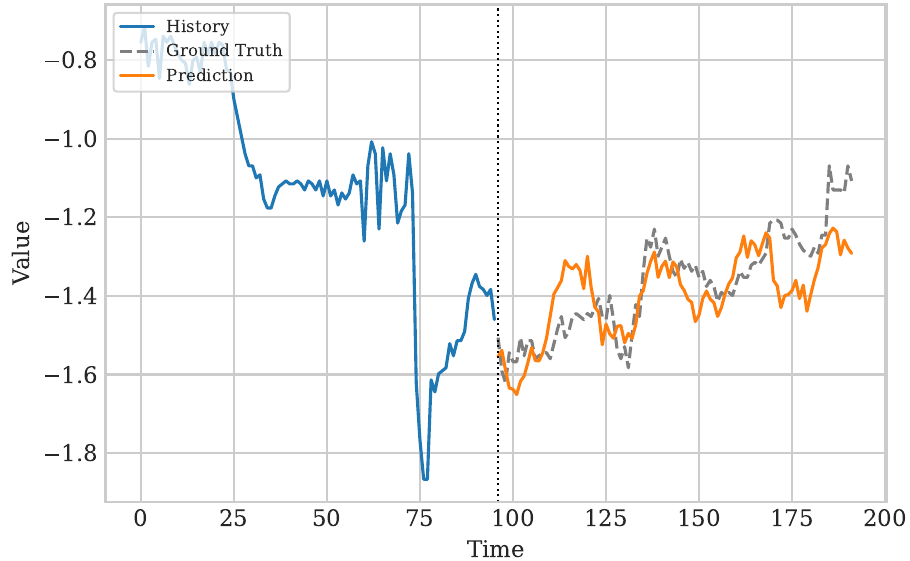}
    \subcaption{Prediction}
\end{subfigure}
\begin{subfigure}{0.31\textwidth}
    \centering
    \includegraphics[width=1\textwidth]{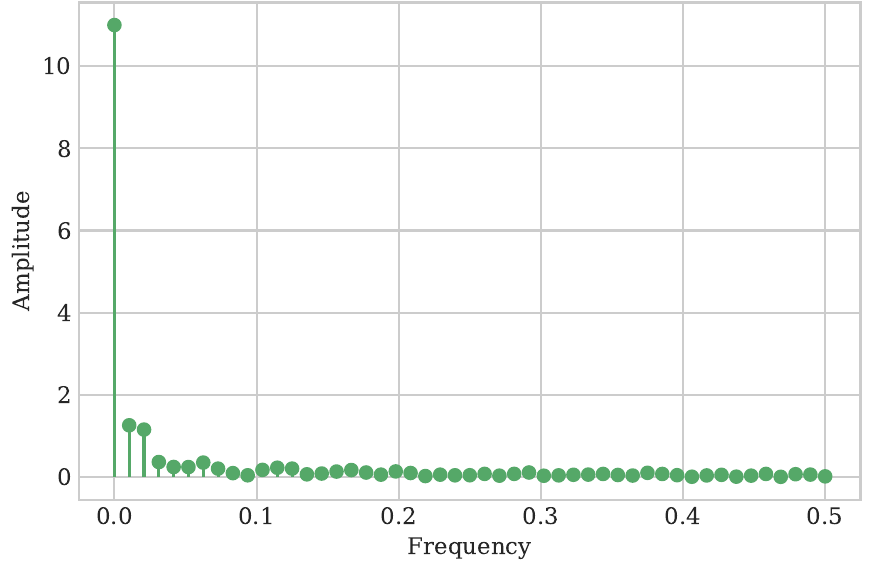}
    \subcaption{Frequency of lookback window}
\end{subfigure}
\begin{subfigure}{0.32\textwidth}
    \centering
    \includegraphics[width=1\textwidth]{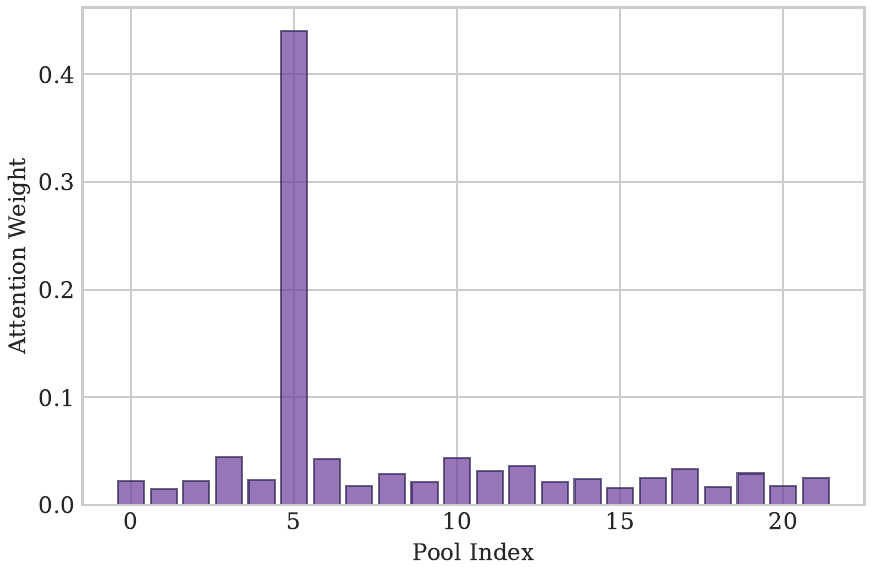}
    \subcaption{Weight of  Pool}
\end{subfigure}
\caption{ETTh1 frequency pool analysis case1.}
\label{fig7}
\vspace{-2mm}
\end{figure*}

\begin{figure*}[h]
\centering
\begin{subfigure}{0.32\textwidth}
    \centering
    \includegraphics[width=1\textwidth]{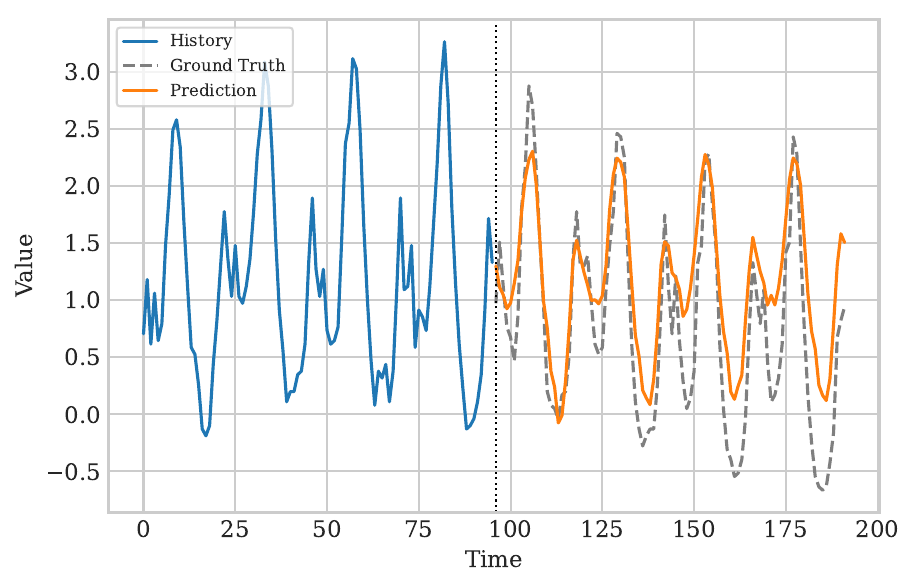}
    \subcaption{Prediction}
\end{subfigure}
\begin{subfigure}{0.32\textwidth}
    \centering
    \includegraphics[width=1\textwidth]{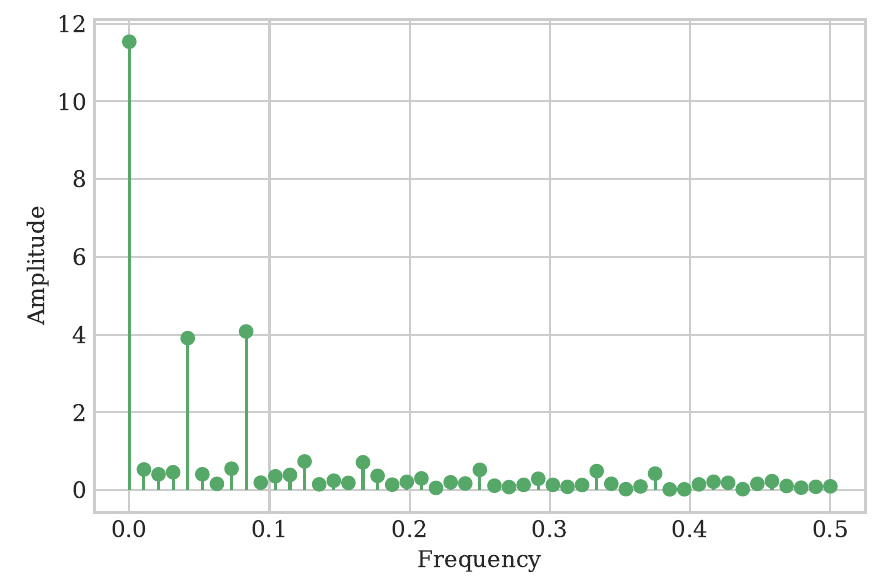}
    \subcaption{Frequency of lookback window}
\end{subfigure}
\begin{subfigure}{0.32\textwidth}
    \centering
    \includegraphics[width=1\textwidth]{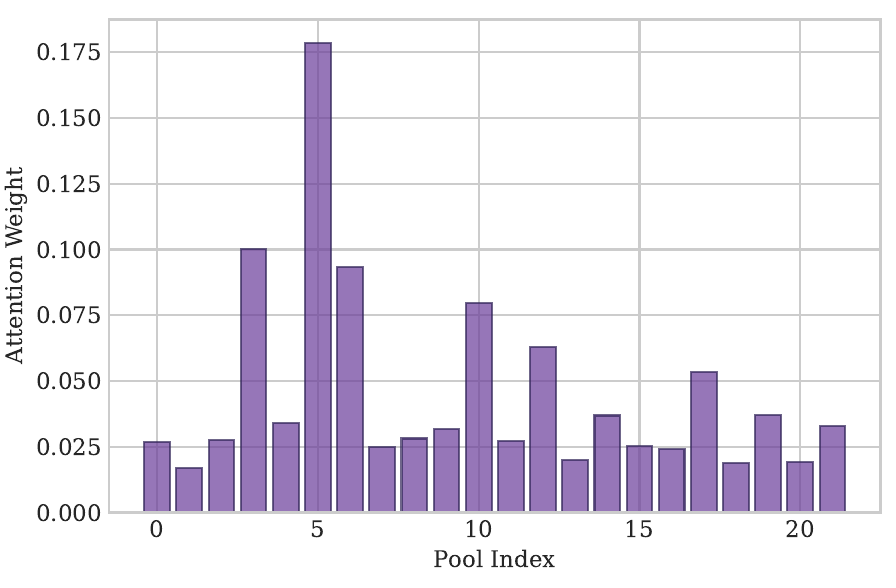}
    \subcaption{Weight of Frequency Pool}
\end{subfigure}
\caption{ETTh1 frequency pool analysis case2.}
\label{fig8}
\vspace{-2mm}
\end{figure*}



\paragraph{Analysis of Frequency Pool} 
To better understand the behavior of the Frequency Pool, we provide case studies in Figures ~\ref{fig7} and ~\ref{fig8}. Subfigure (a) visualizes the forecasting results. Subfigure (b) shows the frequency domain distribution via FFT, while subfigure (c) displays the corresponding Frequency Pool weights. In Figure ~\ref{fig7}(b), a prominent low-frequency peak aligns with a dominant weight in Figure ~\ref{fig7}(c). Similarly, Figure ~\ref{fig8} shows multiple high-frequency components that correlate with specific weight distributions in Figure ~\ref{fig8}(c). These observations support the effectiveness of the FFT-driven indexing mechanism. Furthermore, the results suggest that the Frequency Pool learns relatively decoupled frequency-domain prototypes.

\begin{figure*}[!h]
\centering
\includegraphics[width=0.5\textwidth]{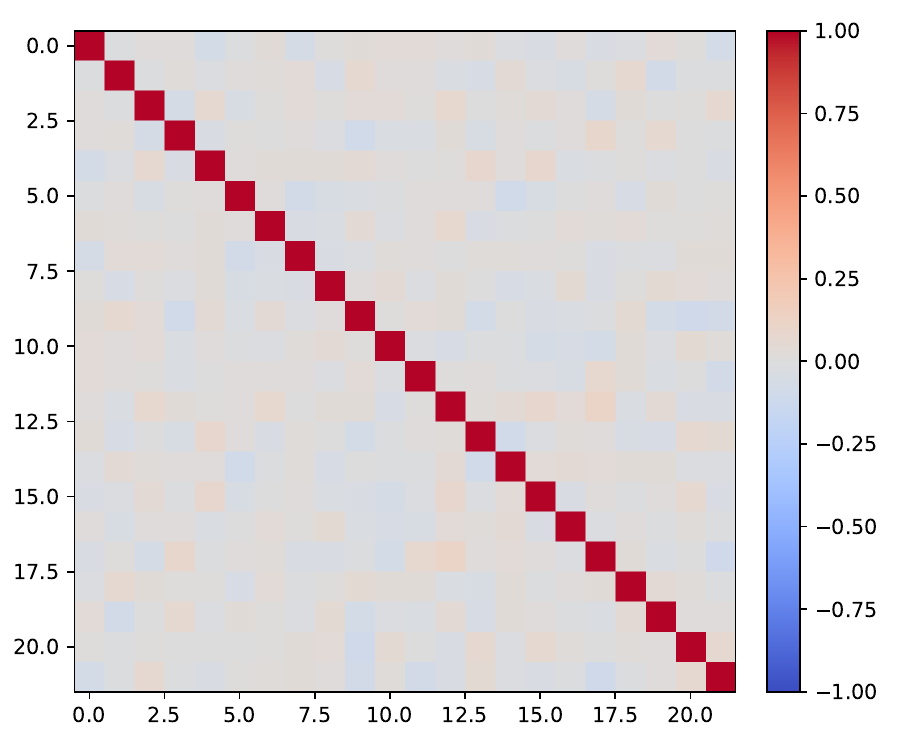}
\caption{Frequency pool Gram matrix.}
\label{fig6}
\end{figure*}

We visualize the cosine similarity matrix of the ETTh1 frequency pool to investigate its representation space. As shown in Figure ~\ref{fig6}, the matrix exhibits a distinct near-identity structure, where diagonal elements approach 1 and off-diagonal entries remain near 0. This high degree of orthogonality indicates that each prompt within the pool captures unique frequency-domain prototypes. 



\begin{figure*}[!h]
\centering
\includegraphics[width=0.5\textwidth]{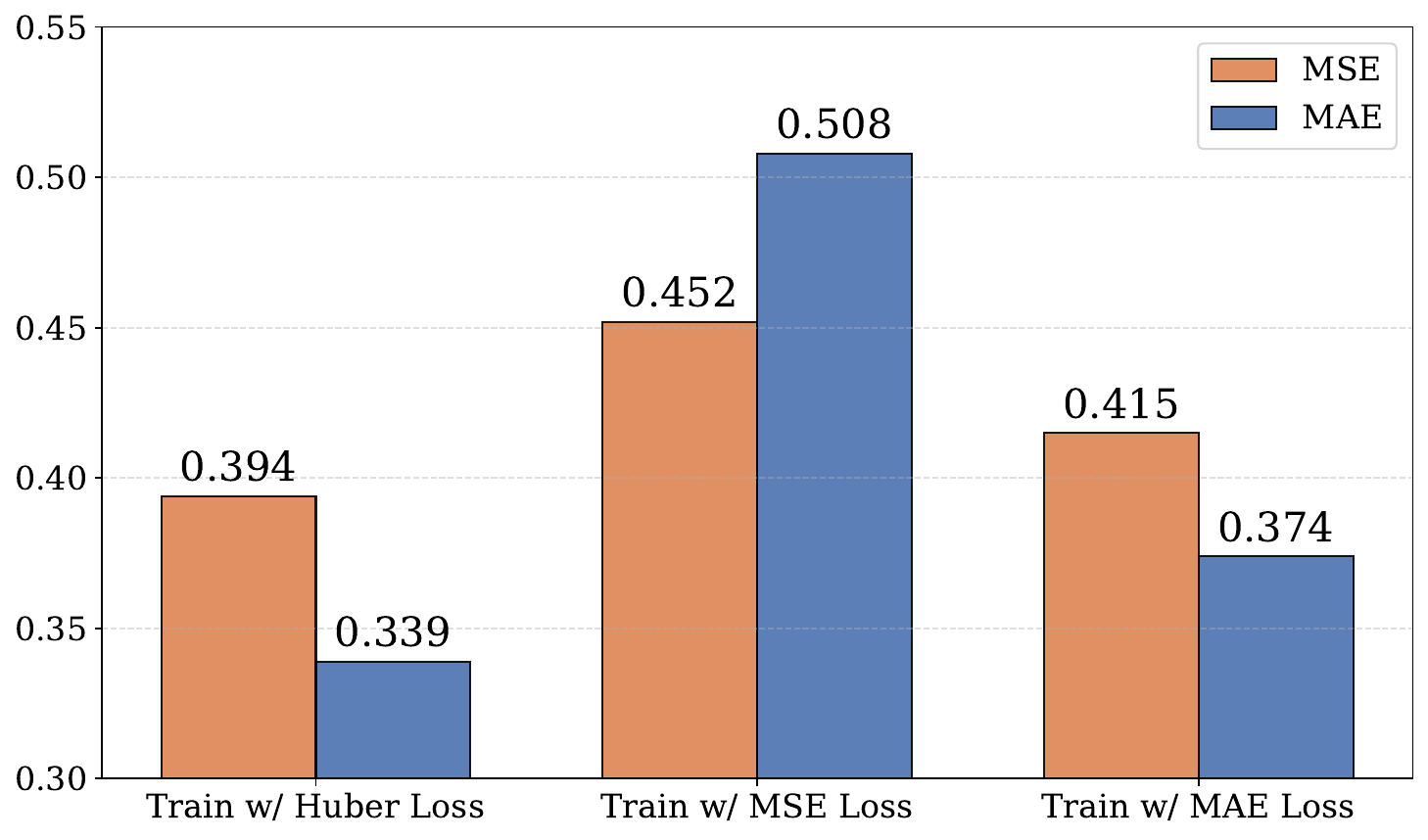}
\caption{Comparison with different losses.}
\label{fig10}
\end{figure*}

\paragraph{Impact of Loss Functions} To evaluate the influence of the optimization objective, we compare the performance of our model using Huber loss against two standard benchmarks: Mean Squared Error (MSE) and Mean Absolute Error (MAE). Figure~\ref{fig10} shows that using the Huber loss leads to better performance than MSE and MAE losses. This suggests that the Huber loss provides a better balance between sensitivity to small errors and robustness to large residuals.

\paragraph{Efficiency Analysis}
Table~\ref{tab:efficiency_analysis} reports the computational efficiency of FreSia and representative baselines on the ECL dataset with a prediction horizon of 336 and a batch size of 8. The reported test time is measured after prompt embeddings are precomputed and saved. Under the cached-prompt evaluation setting, FreSia achieves efficient inference with 11.29 seconds of test time. This is faster than TimeCMA, TimeKAN, iTransformer, and OFA. Although FreSia adopts a frozen LLM as the semantic encoder, the prompt embeddings are precomputed and cached before forecasting. Therefore, the LLM does not need to be repeatedly invoked during online testing, making the additional semantic extraction cost amortized.

In terms of memory and parameters, FreSia uses 2936 MiB GPU memory and 28.29M parameters, which is comparable to TimeCMA with 29.53M parameters and 2690 MiB memory. OFA has fewer parameters and lower GPU memory, but its test time is substantially higher in our evaluation. These results show that FreSia introduces manageable computational overhead while maintaining efficient inference.

\begingroup
\renewcommand{\arraystretch}{0.85}
\begin{table}[!h]
\centering
\small
\setlength{\tabcolsep}{4pt}{
\begin{tabular}{c|ccc}
\toprule
\textbf{Model} & \textbf{Test Time (s)} & \textbf{GPU Memory (MiB)} & \textbf{Params (M)} \\
\midrule
TimeCMA      & 16.78 & 2690 & 29.53 \\
OFA          & 1697.66 & \textbf{916} & 3.92 \\
iTransformer & 129.98 & 2248 & 4.96 \\
TimeKAN      & 44.32 & 3556 & 0.044 \\
FreSia       & \textbf{11.29} & 2936 & 28.29 \\
\bottomrule
\end{tabular}
}
\caption{Efficiency comparison. We report test time, peak GPU memory usage, and model parameters under the same evaluation setting.}
\label{tab:efficiency_analysis}
\end{table}
\endgroup

Furthermore, we evaluate the computational overhead by comparing the token consumption of different prompting strategies. The TimeCMA prompt template contains approximately 226 tokens, whereas the FreSia prompt contains approximately 32 tokens under the same tokenization setting, corresponding to about a 7.1$\times$ reduction in prompt length.


\section{Conclusion}

In this paper, we present an LLM-enhanced framework for multivariate time-series forecasting and anomaly detection. FreSia converts frequency-domain structures into semantic prompts and uses CLS-driven calibration to align these prompts with temporal representations. Experiments on eight forecasting benchmarks and five anomaly-detection benchmarks show that the resulting frequency-semantic representations are effective across both tasks.

\section{Authorship contribution statement}
Yubo Wang: Conceptualization, Methodology, Validation, Visualization,
Writing– original draft, Writing- review \& editing. Hui He: Conceptualization, Investigation, Writing- review \& editing, Supervision. Hezhe Qiao: Conceptualization, Investigation, Writing- review \& editing. Guoqing Ji: Investigation, Writing- review. Zhendong Niu: Conceptualization, Investigation, Writing- review \& editing, Supervision.


\section{Declaration of generative AI and AI-assisted technologies in the manuscript preparation process}

During the preparation of this work the author(s) used ChatGPT and Gemini in order to translate and polish this paper. After using this tool/service, the author(s) reviewed and edited the content as needed and take(s) full responsibility for the content of the published article.

\clearpage
\appendix

\section{Energy-preserving property of DFT} 
\label{app:energy-preserving}

\begingroup

The theoretical motivation for selecting dominant frequency components can be further explained by the energy-preserving property of the discrete Fourier transform (DFT). Let $\mathbf{x}\in\mathbb{R}^{N}$ be a real-valued time series, where $n=0,\dots,N-1$ denotes the time index. Its DFT is denoted as $\mathbf{X}=\mathcal{F}(\mathbf{x})$, where $X[k]$ is the complex Fourier coefficient at the $k$-th discrete frequency bin, and $k=0,\dots,N-1$ denotes the frequency-bin index. Under the standard DFT convention,
\begin{equation}
    X[k]=\sum_{n=0}^{N-1}x[n]\exp\left(-j\frac{2\pi kn}{N}\right),
\end{equation}
where $j$ is the imaginary unit. The zero-frequency coefficient $X[0]$ is the DC component, which corresponds to the sum of the sequence values and reflects the mean level of the sequence after normalization by $N$.

According to Parseval's theorem~\cite{oppenheim1999discrete}, the signal energy is preserved between the time and frequency domains:
\begin{equation}
    \sum_{n=0}^{N-1}|x[n]|^2
    =
    \frac{1}{N}\sum_{k=0}^{N-1}|X[k]|^2 .
\end{equation}
This relation indicates that the magnitude spectrum provides a principled decomposition of signal energy over frequency components.

Following the implementation used in our spectral sparsity analysis, the retained energy ratio is computed on the one-sided spectrum returned by the real-valued FFT. The DC component is excluded because it mainly reflects the mean level rather than periodic variations. Let
\begin{equation}
    \mathcal{I}=\{1,\dots,\lfloor N/2\rfloor\}
\end{equation}
denote the set of one-sided positive non-DC frequency indices. For each $k\in\mathcal{I}$, its normalized frequency is $f_k=k/N$ cycles per time step, and the corresponding period is $T_k=N/k$ time steps. If the sampling interval is $\Delta t$, then $f_k=k/(N\Delta t)$ and $T_k=N\Delta t/k$.

The frequency indices in $\mathcal{I}$ are ranked by their one-sided spectral energy $|X[k]|^2$. Let $\mathcal{S}_K\subseteq\mathcal{I}$ denote the set of the $K$ frequency indices with the largest $|X[k]|^2$. The retained non-DC spectral energy ratio is defined as
\begin{equation}
    \rho_K =
    \frac{\sum_{k\in\mathcal{S}_K}|X[k]|^2}
    {\sum_{k\in\mathcal{I}}|X[k]|^2}.
\end{equation}
A large $\rho_K$ indicates that a small number of dominant frequency components preserve most of the one-sided non-DC spectral energy. Therefore, selecting Top-$K$ frequency components provides a compact approximation of the main temporal variations in the Fourier basis. This supports the use of dominant frequencies as structural primitives for frequency-guided semantic instantiation.

\endgroup

\section{Reconstruction for anomaly detection}
\setcounter{table}{0}
\label{app:anomaly_score}

\begingroup
\renewcommand{\arraystretch}{0.75}
\begin{table*}[!h]
\centering
\small
\setlength{\tabcolsep}{5pt}{
\begin{tabular}{c|ccc}
\toprule
\textbf{Dataset}
& \textbf{Time Residual}
& \textbf{Time-Freq. Residual}
& \textbf{Freq. Residual}
\\
\midrule
SMD  & \textbf{0.863} & 0.392 & 0.162 \\
PSM  & \textbf{0.969} & 0.875 & 0.781 \\
SWaT & \textbf{0.943} & 0.913 & 0.846 \\
MSL  & \textbf{0.856} & 0.171 & 0.080 \\
SMAP & \textbf{0.787} & 0.641 & 0.637 \\
\midrule
Avg. & \textbf{0.884} & 0.598 & 0.501 \\
\bottomrule
\end{tabular}
}
\caption{F1 comparison of anomaly scoring rules. Time Residual denotes the standard reconstruction-error score used by FreSia. Time-Freq. Residual combines time-domain and frequency-domain residuals with equal weight. Freq. Residual uses only the frequency residual.}
\label{tab:appendix_anomaly_score_diagnostic}
\end{table*}
\endgroup


\textbf{Time Residual.} The standard time-domain score computes the point-wise reconstruction error by averaging over variables: \( s_t(i)=\frac{1}{C}\sum_{c=1}^{C}\left(x_{i,c}-\hat{x}_{i,c}\right)^2, \) where $i$ denotes the time index. This score preserves temporal localization because each timestamp has its own anomaly score. 

\textbf{Freq. Residual.} For the frequency-domain score, FFT is applied along the temporal dimension of each window. The DC component is removed to avoid domination by the mean level. The frequency-domain residual is then computed as the average magnitude difference between the spectra of the original and reconstructed windows: 
\begin{equation}
    s_f^{\mathrm{win}} = \frac{1}{FC} \sum_{k=1}^{F} \sum_{c=1}^{C} \left| \left|\mathcal{F}(\mathbf{x})_{k,c}\right| - \left|\mathcal{F}(\hat{\mathbf{x}})_{k,c}\right| \right|,
\end{equation} 
{where $F$ denotes the number of non-DC frequency bins. Since this frequency residual is computed at the window level, it is broadcast to all timestamps within the window: \( s_f(i)=s_f^{\mathrm{win}}, \quad i=1,\dots,L. \) Thus, Freq. Residual captures window-level spectral mismatch but does not provide point-wise temporal localization. }

\textbf{Time-Freq. Residual.} To avoid scale mismatch between the time-domain and frequency-domain scores, both scores are normalized using the mean and standard deviation computed from the training set: \( \tilde{s}_t=\frac{s_t-\mu_t}{\sigma_t+\epsilon}, \qquad \tilde{s}_f=\frac{s_f-\mu_f}{\sigma_f+\epsilon}. \) The combined score is then defined as \( s_{\mathrm{tf}}=\alpha \tilde{s}_t+(1-\alpha)\tilde{s}_f, \) where $\alpha=0.5$ is used in our diagnostic experiment. 


Table~\ref{tab:appendix_anomaly_score_diagnostic} shows that the standard Time Residual achieves the best F1 score on all five datasets, with the highest average F1 of 0.884. In contrast, Time-Freq. Residual and Freq. Residual obtain much lower average F1 scores of 0.598 and 0.501, respectively. This suggests that directly adding a frequency-domain residual is not a robust scoring strategy for anomaly detection in this setting. One possible reason is that the frequency residual measures window-level spectral mismatch and is assigned to all timestamps in the window, which weakens point-wise anomaly localization. Therefore, the standard reconstruction-error score is retained, and anomaly detection is used to evaluate representation generalization rather than to claim a new anomaly-specific scoring rule.

\renewcommand{\thetable}{\Alph{section}.\arabic{table}}
\setcounter{table}{0}

\section{Comparison with Frequency-domain Models}
\setcounter{table}{0}
\label{app:frequency baseline}

\begingroup
\renewcommand{\arraystretch}{0.5}
\begin{table*}[!h]
\centering
\setlength{\tabcolsep}{1mm}{
\begin{tabular}{cc|cc|cc|cc|cc|cc}
\toprule
\multicolumn{2}{c|}{\textbf{Models}}
& \multicolumn{2}{c|}{\textbf{FreSia}}
& \multicolumn{2}{c|}{TimeKAN}
& \multicolumn{2}{c|}{PaiFilter}
& \multicolumn{2}{c|}{TexFilter}
& \multicolumn{2}{c}{FredFormer}
\\
\cmidrule(r){1-2}\cmidrule(r){3-4} \cmidrule(r){5-6} \cmidrule(r){7-8}\cmidrule(r){9-10} \cmidrule(r){11-12}
\multicolumn{2}{c|}{\textbf{Metrics}}
& \textbf{MSE} & \textbf{MAE}
& \textbf{MSE} & \textbf{MAE}
& \textbf{MSE} & \textbf{MAE}
& \textbf{MSE} & \textbf{MAE}
& \textbf{MSE} & \textbf{MAE}
\\
\midrule

\multirow{5}{*}{\rotatebox{90}{ETTm1}}
& 96 &\textcolor{red}{0.310} &\textcolor{red}{0.348} &0.322 &0.361 &\textcolor{blue}{0.318} &\textcolor{blue}{0.358} &0.321 &0.361 &0.328 &0.363 \\
& 192 &\textcolor{red}{0.352} &\textcolor{red}{0.374} &\textcolor{blue}{0.357} &0.383 &0.364 &0.383 &0.367 &0.387 &0.367 &\textcolor{blue}{0.382} \\
& 336 &\textcolor{red}{0.378} &\textcolor{red}{0.396} &\textcolor{blue}{0.383} &\textcolor{blue}{0.402} &0.396 &0.406 &0.401 &0.409 &0.395 &0.403 \\
& 720 &\textcolor{red}{0.429} &\textcolor{red}{0.427} &\textcolor{blue}{0.446} &\textcolor{blue}{0.436} &0.456 &0.444 &0.477 &0.448 &0.454 &0.440 \\
& Avg &\textcolor{red}{0.367} &\textcolor{red}{0.386} &\textcolor{blue}{0.377} &\textcolor{blue}{0.395} &0.384 &0.398 &0.392 &0.401 &0.386 &0.397 \\

\midrule
\multirow{5}{*}{\rotatebox{90}{ETTm2}}
& 96 &\textcolor{red}{0.169} &\textcolor{red}{0.251} &\textcolor{blue}{0.174} &\textcolor{blue}{0.255} &0.174 &0.257 &0.175 &0.258 &0.178 &0.261 \\
& 192 &\textcolor{red}{0.237} &\textcolor{red}{0.296} &\textcolor{blue}{0.239} &\textcolor{blue}{0.299} &0.240 &0.300 &0.240 &0.301 &0.244 &0.303 \\
& 336 &\textcolor{blue}{0.298} &\textcolor{red}{0.337} &0.301 &0.340 &\textcolor{red}{0.297} &\textcolor{blue}{0.339} &0.311 &0.347 &0.302 &0.341 \\
& 720 &\textcolor{red}{0.392} &\textcolor{blue}{0.395} &0.395 &0.396 &\textcolor{blue}{0.392} &\textcolor{red}{0.393} &0.414 &0.405 &0.397 &0.396 \\
& Avg &\textcolor{red}{0.274} &\textcolor{red}{0.320} &0.277 &0.323 &\textcolor{blue}{0.276} &\textcolor{blue}{0.322} &0.285 &0.328 &0.280 &0.325 \\

\midrule
\multirow{5}{*}{\rotatebox{90}{ETTh1}}
& 96 &\textcolor{red}{0.361} &\textcolor{red}{0.389} &\textcolor{blue}{0.367} &0.395 &0.375 &\textcolor{blue}{0.394} &0.382 &0.402 &0.376 &0.394 \\
& 192 &\textcolor{red}{0.406} &\textcolor{red}{0.418} &\textcolor{blue}{0.414} &\textcolor{blue}{0.420} &0.436 &0.422 &0.430 &0.429 &0.440 &0.426 \\
& 336 &\textcolor{red}{0.419} &\textcolor{red}{0.430} &\textcolor{blue}{0.445} &\textcolor{blue}{0.434} &0.476 &0.443 &0.472 &0.451 &0.472 &0.440 \\
& 720 &\textcolor{blue}{0.446} &\textcolor{red}{0.455} &\textcolor{red}{0.444} &\textcolor{blue}{0.459} &0.474 &0.469 &0.481 &0.473 &0.490 &0.466 \\
& Avg &\textcolor{red}{0.408} &\textcolor{red}{0.423} &\textcolor{blue}{0.417} &\textcolor{blue}{0.427} &0.440 &0.432 &0.441 &0.439 &0.444 &0.432 \\

\midrule
\multirow{5}{*}{\rotatebox{90}{ETTh2}}
& 96 &\textcolor{red}{0.270} &\textcolor{red}{0.328} &\textcolor{blue}{0.290} &\textcolor{blue}{0.340} &0.292 &0.343 &0.293 &0.343 &0.293 &0.343 \\
& 192 &\textcolor{red}{0.346} &\textcolor{red}{0.380} &0.375 &0.392 &\textcolor{blue}{0.369} &0.395 &0.374 &0.396 &0.370 &\textcolor{blue}{0.390} \\
& 336 &\textcolor{red}{0.355} &\textcolor{red}{0.394} &0.423 &0.435 &0.420 &0.432 &0.417 &0.430 &\textcolor{blue}{0.385} &\textcolor{blue}{0.413} \\
& 720 &\textcolor{red}{0.400} &\textcolor{red}{0.428} &0.443 &0.449 &0.430 &0.446 &0.449 &0.460 &\textcolor{blue}{0.419} &\textcolor{blue}{0.439} \\
& Avg &\textcolor{red}{0.343} &\textcolor{red}{0.382} &0.383 &0.404 &0.378 &0.404 &0.383 &0.407 &\textcolor{blue}{0.367} &\textcolor{blue}{0.396} \\

\midrule
\multirow{5}{*}{\rotatebox{90}{ECL}}
& 96 &\textcolor{red}{0.145} &\textcolor{blue}{0.243} &0.174 &0.266 &0.176 &0.264 &\textcolor{blue}{0.147} &0.245 &0.148 &\textcolor{red}{0.242} \\
& 192 &\textcolor{blue}{0.163} &0.260 &0.182 &0.273 &0.185 &0.270 &\textcolor{red}{0.160} &\textcolor{red}{0.250} &0.164 &\textcolor{blue}{0.258} \\
& 336 &0.180 &0.276 &0.197 &0.286 &0.202 &0.286 &\textcolor{red}{0.173} &\textcolor{red}{0.267} &\textcolor{blue}{0.178} &\textcolor{blue}{0.274} \\
& 720 &\textcolor{blue}{0.212} &\textcolor{blue}{0.306} &0.236 &0.320 &0.242 &0.319 &\textcolor{red}{0.210} &0.309 &0.213 &\textcolor{red}{0.301} \\
& Avg &\textcolor{blue}{0.175} &0.271 &0.197 &0.286 &0.201 &0.285 &\textcolor{red}{0.172} &\textcolor{red}{0.268} &0.176 &\textcolor{blue}{0.269} \\

\midrule
\multirow{5}{*}{\rotatebox{90}{Weather}}
& 96 &\textcolor{red}{0.150} &\textcolor{red}{0.192} &\textcolor{blue}{0.162} &0.208 &0.164 &0.210 &0.162 &\textcolor{blue}{0.207} &0.163 &0.207 \\
& 192 &\textcolor{red}{0.202} &\textcolor{red}{0.241} &\textcolor{blue}{0.207} &\textcolor{blue}{0.249} &0.214 &0.252 &0.210 &0.250 &0.209 &0.250 \\
& 336 &\textcolor{red}{0.259} &\textcolor{red}{0.284} &\textcolor{blue}{0.263} &\textcolor{blue}{0.290} &0.268 &0.293 &0.265 &0.290 &0.265 &0.292 \\
& 720 &\textcolor{red}{0.336} &\textcolor{red}{0.335} &\textcolor{blue}{0.338} &0.341 &0.344 &0.342 &0.342 &\textcolor{blue}{0.340} &0.342 &0.341 \\
& Avg &\textcolor{red}{0.236} &\textcolor{red}{0.263} &\textcolor{blue}{0.242} &0.272 &0.247 &0.274 &0.245 &\textcolor{blue}{0.272} &0.245 &0.272 \\

\midrule
\multirow{5}{*}{\rotatebox{90}{Exchange}}
& 96 &\textcolor{red}{0.081} &\textcolor{red}{0.200} &0.087 &0.206 &\textcolor{blue}{0.083} &\textcolor{blue}{0.202} &0.091 &0.211 &0.089 &0.208 \\
& 192 &\textcolor{blue}{0.177} &\textcolor{blue}{0.299} &0.181 &0.299 &\textcolor{red}{0.174} &\textcolor{red}{0.296} &0.186 &0.305 &0.181 &0.301 \\
& 336 &\textcolor{red}{0.318} &\textcolor{red}{0.407} &0.347 &0.426 &\textcolor{blue}{0.326} &\textcolor{blue}{0.413} &0.380 &0.449 &0.352 &0.426 \\
& 720 &\textcolor{red}{0.781} &\textcolor{red}{0.668} &0.995 &0.748 &\textcolor{blue}{0.840} &\textcolor{blue}{0.670} &0.896 &0.712 &0.988 &0.736 \\
& Avg &\textcolor{red}{0.339} &\textcolor{red}{0.394} &0.402 &0.420 &\textcolor{blue}{0.356} &\textcolor{blue}{0.395} &0.388 &0.419 &0.402 &0.418 \\

\midrule
\multirow{5}{*}{\rotatebox{90}{ILI}}
& 24 &\textcolor{red}{1.581} &\textcolor{red}{0.792} &2.392 &0.945 &2.500 &0.970 &\textcolor{blue}{2.095} &\textcolor{blue}{0.852} &2.379 &0.947 \\
& 36 &\textcolor{red}{1.373} &\textcolor{red}{0.717} &\textcolor{blue}{1.935} &\textcolor{blue}{0.879} &2.425 &1.015 &2.099 &0.885 &2.323 &0.954 \\
& 48 &\textcolor{red}{1.730} &\textcolor{red}{0.803} &2.244 &0.935 &2.551 &1.028 &\textcolor{blue}{2.075} &\textcolor{blue}{0.880} &2.657 &1.070 \\
& 60 &\textcolor{red}{1.568} &\textcolor{red}{0.802} &2.090 &0.960 &2.582 &1.038 &\textcolor{blue}{2.070} &\textcolor{blue}{0.902} &2.994 &1.140 \\
& Avg &\textcolor{red}{1.563} &\textcolor{red}{0.779} &2.165 &0.930 &2.514 &1.013 &\textcolor{blue}{2.085} &\textcolor{blue}{0.880} &2.588 &1.028 \\

\bottomrule
\end{tabular}
}

\caption{Comparison with frequency-domain baselines. The input sequence length is 36 for the ILI dataset and 96 for others. The best results are in red and the second best are blue.}
\label{tab:appendix_frequency_baselines}
\end{table*}
\endgroup


Table~\ref{tab:appendix_frequency_baselines} compares FreSia with representative frequency-domain forecasting models. Although these baselines also exploit spectral information, they mainly model frequency-domain patterns through numerical transformations or filtering. In contrast, FreSia further converts dominant frequency components into semantic cycle descriptors and aligns them with LLM representations. FreSia achieves the best average performance on 7 out of 8 datasets and ranks first in 14 out of 16 dataset-level average metrics. This shows that the gains of FreSia cannot be attributed merely to the use of frequency-domain information. Instead, the results suggest that frequency-semantic instantiation provides additional benefits over purely numerical frequency-domain modeling. TexFilter performs slightly better on ECL, indicating that frequency filtering remains effective in some settings; nevertheless, FreSia shows stronger overall performance across diverse benchmarks.

\section{Robustness to Auxiliary Trend / Volatility Mappings}
\setcounter{table}{0}
\label{app:robustness mappings}

\renewcommand{\arraystretch}{0.85}
\begin{table}[t]
\centering
\small
\setlength{\tabcolsep}{6pt}
\begin{tabular}{c|cc|cc|cc}
\toprule
\multirow{2}{*}{Variant} 
& \multicolumn{2}{c|}{ETTm2} 
& \multicolumn{2}{c|}{ECL} 
& \multicolumn{2}{c}{Avg.} \\
\cmidrule(lr){2-3} \cmidrule(lr){4-5} \cmidrule(lr){6-7}
& MSE & MAE & MSE & MAE & MSE & MAE \\
\midrule
Adaptive-Quantile 
& 0.276 & 0.321 & 0.177 & 0.274 & 0.227 & 0.298 \\
Continuous        
& 0.274 & 0.320 & 0.178 & 0.274 & 0.226 & 0.297 \\
Freq-only         
& 0.276 & 0.322 & 0.179 & 0.276 & 0.228 & 0.299 \\
\textbf{FreSia}            
& \textbf{0.274} & \textbf{0.320} & \textbf{0.175} & \textbf{0.271} & \textbf{0.224} & \textbf{0.296} \\
\bottomrule
\end{tabular}
\caption{Robustness to auxiliary trend/volatility mappings. The results are averaged over all horizons.}
\label{tab:tv_mapping_robustness}
\end{table}

Table~\ref{tab:tv_mapping_robustness} evaluates whether FreSia relies on the fixed trend and volatility thresholds in the default FGPrompt mapping rules. The adaptive-quantile variant replaces fixed thresholds with quantile-based thresholds estimated from the training set, while the continuous variant removes discrete trend/volatility labels and directly uses the corresponding continuous scores. These alternatives achieve comparable results to the default mapping, indicating that FreSia is not sensitive to a specific threshold design. Moreover, the freq-only variant removes all trend and volatility descriptors, thereby completely eliminating these hand-crafted thresholds, but still remains close to the full model. These results show that the hard-coded trend/volatility mappings are not the core source of FreSia’s performance; the main structural cues come from frequency-derived cycle descriptors.

\clearpage







\bibliographystyle{elsarticle-num-names}
\bibliography{cite_new}



\end{document}